# Vox Populi, Vox AI? Using Language Models to Estimate German Public Opinion

Leah von der Heyde [1,2] *, Anna-Carolina Haensch [1, 3], Alexander Wenz [4]

[1] *LMU Munich*, [2] *Munich Center for Machine Learning*, [3] *University of Maryland*, [4] *University of Mannheim, Mannheim Centre for European Social Research*

* *Corresponding author:* L.Heyde@lmu.de

## Abstract

The recent development of large language models (LLMs) has spurred discussions about whether LLM-generated "synthetic samples" could complement or replace traditional surveys, considering their training data potentially reflects attitudes and behaviors prevalent in the population. A number of mostly US-based studies have prompted LLMs to mimic survey respondents, with some of them finding that the responses closely match the survey data. However, several contextual factors related to the relationship between the respective target population and LLM training data might affect the generalizability of such findings. In this study, we investigate the extent to which LLMs can estimate public opinion in Germany, using the example of vote choice. We generate a synthetic sample of personas matching the individual characteristics of the 2017 German Longitudinal Election Study respondents. We ask the LLM GPT-3.5 to predict each respondent's vote choice and compare these predictions to the survey-based estimates on the aggregate and subgroup levels. We find that GPT-3.5 does not predict citizens' vote choice accurately, exhibiting a bias towards the Green and Left parties. While the LLM captures the tendencies of "typical" voter subgroups, such as partisans, it misses the multifaceted factors swaying individual voter choices. By examining the LLM-based prediction of voting behavior in a new context, our study contributes to the growing body of research about the conditions under which LLMs can be leveraged for studying public opinion. The findings point to disparities in opinion representation in LLMs and underscore the limitations in applying them for public opinion estimation.

**Keywords:** LLM, artificial intelligence, survey data, synthetic samples, voting behavior, public opinion

## 1. Introduction

The recent development and large-scale proliferation of large language models (LLMs), such as OpenAI's GPT (OpenAI et al., 2023) or Meta's Llama (Touvron et al., 2023), have spurred discussions about the extent to which these language models can be used for research in the social and behavioral sciences. Researchers have started to explore various applications to facilitate the collection and analysis of survey data. Examples include the use of LLMs for questionnaire design and scale development (Götz et al., 2023; Hernandez & Nie, 2022; Konstantis et al., 2023; Laverghetta & Licato, 2023, Lee et al., 2023), conducting interviews (Chopra & Haaland, 2023; Cuevas Villalba et al., 2023), coding open-ended survey responses (Mellon et al., 2023; Rytting et al., 2023), imputing missing data and detecting statistical outliers (Jaimovitch-López et al., 2023; Kim & Lee, 2023), detecting non-human respondents in online surveys (Lebrun et al., 2023), and data visualization and interpretation (Liew & Mueller, 2022; Sultanum & Srinivasan, 2023).

Beyond augmenting survey data collection and analysis, research has also started to examine to what extent LLMs can be used for making valid inferences about a population (e.g. Argyle et al., 2023). LLMs are trained on large amounts of internet text data, such as selected book collections, Wikipedia, and social media data, which potentially reflect attitudes and behaviors prevalent in the population. Their text output to a request represents a conditional probability based on the training data and the specific contextual information provided in the request. Thus, LLMs might serve as a novel method of collecting data about public opinion. *Synthetic samples* generated by LLMs might be particularly useful for collecting data faster and at lower cost compared to surveys and might allow for covering different population segments, including those that are potentially hard to reach with surveys. Such samples can be created by sequentially feeding individual socio-demographic, socio-economic, and/or attitudinal information of specific persons to the LLM and asking it to respond to survey questions from the respective person's perspective.

While there has been an increasing number of studies about the use of LLMs for population inference, most existing research has focused on the United States. We argue that the generalizability of such findings beyond the US population is questionable because the suitability of LLMs for estimating public opinion depends on a variety of contextual factors associated with the target population. These factors include (1) the prevalence of native-language training data, (2) a country's political and societal structure, which has a complex relationship with public opinion that can vary across countries and might not be equally reflected in the training data, as well as (3) structural differences between the target population and the population reflected in the training data. Polling voting behavior is one relevant and much-researched example of public opinion estimation. It is also an example that is heavily dependent on the national social and political context. For example, the dynamics of vote choice are markedly different in a multi-party system, such as Germany's, than in the US two-party system. At the same time, due to its linguistic and socio-demographic presence online and its socio-political structure, Germany presents a reasonable middle ground for the examination of LLM public opinion estimation, the results of which can be telling for societies represented in LLM training data even less.

In this paper, we examine to what extent LLMs can estimate public opinion in Germany by addressing the following research questions:

**RQ1.** Do LLM-based samples provide similar estimates of voting behavior as national election studies?

**RQ2.** How do LLMs' estimates of voting behavior deviate from national election studies for different subgroups of the population?

Following the approach employed by Argyle et al. (2023), we create a synthetic sample of eligible voters based on data from the German Longitudinal Election Study (GLES). These personas include individual-level information on variables that in the literature have been found to be important predictors of voting behavior – demographics, party affiliations, and views on politically salient issues, such as immigration. Based on this information, we prompt the LLM GPT-3.5 to predict the voting behavior of each individual. From the LLM responses, we extract the predicted vote choices for each persona and compare them to the voting behavior reported by respondents in the GLES data. Thus, our primary goal in this paper is not to assess whether LLMs can predict actual election outcomes, but whether they can infer individual voting behavior and arrive at estimates comparable to those made with individual-level survey data.

Using the example of voting behavior, we provide a twofold methodological contribution to public opinion estimation using LLMs. We (1) show how a popular LLM performs in estimating voting behavior in Germany compared to survey data, and (2) analyze which individual-level factors influence its predictions. Overall, in investigating the suitability of using LLMs for public opinion estimation in a new context, our study contributes to the growing body of research on the extent to which LLMs can be leveraged for research in the social sciences.

## 2. Background

In survey research, synthesizing respondent samples is one especially relevant application of LLMs. Such samples would allow for pre-testing survey questions on different population segments faster and cheaper. They could also potentially supplement or even replace survey-based data collection and public opinion estimation based on human samples, for example, in the context of political polls estimating voting behavior. The underlying idea survey researchers leverage is that LLMs are based on human-created data and could therefore potentially reflect humans' underlying attitudes and behaviors.

Trained on vast amounts of text data, LLMs generate a conditional probability distribution of how likely given tokens, i.e., particles of words, are followed by specific other tokens. Presented with a string of words (*LLM input*), LLMs then draw on this probability distribution to predict words that are likely to follow (*LLM output*). For example, given the input "In the 2020 US presidential elections, I voted for", LLMs are more likely to complete the sentence with "the Democratic candidate" or "the Republican candidate" than with other terms unrelated to candidates or parties. The sentence is more or less likely to be completed with either vote choice depending on the training data, the configuration of the LLM algorithm, as well as any other information provided as input. LLMs are based on large, selected corpora of internet-sourced data, such as selected websites, book collections, and social media data, for example Reddit data from selected subreddits (see e.g. Brown et al., 2020). As this training data includes factual, attitudinal, and behavioral data about people, LLMs might provide a novel method for estimating public opinion in a population by creating synthetic samples: LLMs can be prompted repeatedly to answer survey questions, mimicking human respondents by providing individual-level characteristics as input. The distribution of responses provided in the output could serve as an estimate of the population. However, as of yet, widely-used LLMs do not learn from new data in real-time, but instead are trained on historical data up to a certain time point (see e.g. OpenAI, 2023). Therefore, these LLMs cannot take into account new information on current events that might influence public opinion.

Several recent studies have investigated the potential use of LLMs for replicating or replacing human samples in public opinion research, particularly in the area of political polling. For example, Argyle et al. (2023) prompted GPT-3 to respond to survey questions from the American National Election Study (ANES), reflecting different demographic subgroups of the population. The study found that the LLM-generated responses, on aggregate, closely matched the actual responses in the ANES data, and suggests that LLMs might even be able to estimate public opinion and voting behavior for time points exceeding their own training data. Similarly, Chu et al. (2023) showed that BERT, when trained on news media data, can emulate the attitudes of US subpopulations who consumed news media. Benchmarking the LLM responses against distributions from several surveys by Pew Research Center and the University of Michigan, their findings are robust to prompt wording and variation in media input. Other studies, however, have come to conflicting conclusions. For example, having GPT-3.5 impersonate ANES respondents and answer a set of survey items, the results by Bisbee et al. (2024) were mixed. While the average item scores produced by the LLM were similar to those obtained from the survey data, the LLM-based results had a smaller variance and resulted in different coefficients when regressing the prompt variables on the response. Furthermore, the responses were not robust to prompt wording and across time. Dominguez-Olmedo et al. (2023) had a large range of different language models respond to an entire questionnaire, benchmarking against the American Community Survey. In this study, however, even the aggregate estimates derived from the LLM responses did not match those of the human population. Finally, Santurkar et al. (2023), using the American Trends Panel survey, discovered substantial misalignments for specific subgroups. Testing several LLMs' "default" responses, not providing any further contextual information, as well as responses when prompting the LLMs to impersonate certain subgroups, the authors concluded that LLM-based samples cannot replicate human samples.

A limitation of these existing studies is that they almost exclusively focus on the US population. To better understand the conditions under which LLMs can be used for public opinion research, it is crucial to assess whether they can also be applied for research in other national contexts. Several factors might limit the generalizability of previous findings beyond the United States.

First, it is likely that LLMs are better able to emulate public opinion for the United States than for other countries due to country-level factors associated with the training data. Since LLMs are trained on text data from the internet, the amount of available native-language training data for developing LLMs is considerably smaller for any country with a native language other than English. For example, less than five percent of content on the internet is estimated to be German, compared to English with over 50% (W3Tech, 2023). It is unclear how LLMs transfer their “knowledge” between training data in different languages and what “knowledge” is accessed when prompted in English about a non-English-speaking population (see e.g. Nie et al., 2024a,b, Lai et al., 2023). In either of these two processes, native, potentially more authentic, “knowledge” risks being underrepresented if LLMs are only accessing English-language training data. Moreover, a country’s societal and political structures may differentially affect the determinants of public opinion. These idiosyncratic relationships may not be sufficiently represented in LLM training data. For example, Argyle et al. (2023) showed that GPT-3 mirrored the relationships between subgroup characteristics and voting behavior in the US two-party system. It is unclear, however, whether these findings can be extended to multi-party systems, where the dynamics of voting behavior can follow fundamentally different patterns (Campbell et al., 1960; Lazarsfeld et al., 1944). Predicting voting behavior in multi-party parliamentary democracies is inherently more difficult than predicting the two-party, first-past-the-post presidential democracy of the United States. Statistically, at a very basic level, the probability of making a correct prediction is inversely proportional to the number of parties competing. Moreover, the higher complexity of multi-party-systems, also in terms of more potential combinations of issue positions, make the voting decision more complex for voters. The clear binary alignment of certain issue positions is not obvious outside of the United States: „The next time a Martian visits earth, try to explain to him why those who favor allowing the elimination of a fetus in the mother’s womb also oppose capital punishment. Or try to explain to him why those who accept abortion are supposed to be favorable to high taxation but against a strong military“ (Taleb, 2007, p. 16). Finally, in many multi-party systems, proportional representation and minimum thresholds create voters who vote strategically. These complex decision-making processes are often made spontaneously, in response to parties' popularities in current polls and the specific voting district, and therefore not explicitly discussed online. The concept of “swing voters” therefore is slightly different from that of the United States, as it is simply more common for voters to switch parties depending on the context (regarding policy issues and party popularity) in which the election takes place. Not the least because it is usually the more politically interested and polarized who post on the Internet (e.g. Kim et al., 2021, Tucker et al., 2021, Muhlberger 2003), Internet discussions, however, often have the tendency to conflate political complexities to two camps (e.g. Yarchi et al., 2021). It is therefore likely that LLMs cannot mirror the more complex decision-making process in multi-party systems given the available training data. Additionally, different social structures can lead to different policy-issue salience and conflicts. When inferring from information on demographic or attitudinal subgroups to voting behavior without sufficient “training” in these differences, it thus is likely that LLMs wrongly project the more prominent interest conflicts of the United States onto other contexts.

Second, it is very likely that the training data is affected by coverage bias. The difference between the general population and the population of internet users, the so-called “digital divide” (e.g. Lutz, 2019), may impact how representative the training data is of the population (see e.g. Clemmensen et al., 2023). For example, the socio-demographic digital divide in Germany is slightly different from that in the United States (see Schumacher & Kent, 2020). As the composition of the online and offline populations differs between regions and countries (see also International Telecommunication Union, 2022), a country's societal structure may affect the bias in the LLM training data used to estimate public opinion. In addition, there may be structural and attitudinal differences related to how people in a given society use the internet, that is, between those who actively produce or contribute to the text captured and more passive internet users in general, and between the authors of texts selected for training LLMs and other internet users specifically. For instance, the training data for GPT-3 is not a random sample of internet text, but heavily relies on very few sources, including Wikipedia, Reddit, and two collections of books (Brown et al., 2020) – sources that generally tend to be authored by rather homogenous communities: For example, Wikipedia reports that a plurality (20%) of its editors reside in the United States, edit the English Wikipedia (76%), and that, among editors of the English Wikipedia, 84% are male (Wikipedia, 2023, c.f. Hill & Shaw, 2013). Overall, the "knowledge sources" of LLMs are heavily concentrated on the English-speaking, US context, which are then reflected in their outputs (Johnson et al., 2022).

These factors converge in what can be described as a "black box" of LLMs' internal workings. In this paper, we seek to empirically assess whether or not previous findings regarding public opinion estimation with LLMs can be generalized in the first place, not why they are (not) generalizable, as empirically testing the latter is not only contingent on the former, but would also require a broader scope and insights into the LLM "black box" that the research community does not currently have.

Although there has been some cross-national and cross-lingual research on attitudinal biases of LLMs, these studies either did not explicitly estimate public opinion in general or did not do so for different population subgroups. For example, Motoki et al. (2023) and Hartmann et al. (2023) found that GPT's default political orientation is biased towards left or progressive ideologies in several two- and multi-party systems. Prompting ChatGPT with political questions that can be mapped onto ideological coordinates, Motoki et al. (2023) compared its responses given without any context to those it gave impersonating a partisan and found that the context-less default was more similar to the left partisan. However, the authors did not investigate the individual attitudes of the general public, but instead showcased what GPT "believes" a-priori partisans' political ideology to be (Motoki et al., 2023) or extrapolated from ChatGPT's responses to voting advice application questions to its likely vote choice (Hartmann et al., 2023). Durmus et al.'s (2023) cross-national study is closer to the synthetic-sample approach. The authors tested a custom LLM on entire questionnaires, both its default and when impersonating people from different countries. When comparing the LLM responses to several cross-national survey datasets (Pew Global Attitudes and the World Values Survey), they found that the LLM default responses tended to be more similar to the American and European benchmark data and reflected harmful country-level stereotypes for the other countries. Translations to a country's target language did not always improve the LLM responses' similarity to its speakers' attitudes. But while Durmus et al. (2023) compared English to Russian, Chinese, and Turkish prompting, the authors only used generic country personas ("How would someone from [country] answer this question?"), without considering specific subgroups, allowing only for aggregate cross-country comparisons. Thus, it remains unclear to what extent LLMs can be used for estimating individual-level public opinion outside the much-researched, two-party, English-dominated context of the United States.

In our study, we assess LLMs' suitability for estimating public opinion in Germany by focusing on voting behavior, which is a frequently studied outcome of interest in public opinion research. Germany serves as an example of a Western European democracy, with public opinion formed in the context of not two, but several political parties. Germany has a parliamentary electoral system with proportional representation and its multi-party system is currently characterized by six parties (Schmitt-Beck et al., 2022a): the center-right Christian conservatives (CDU/CSU), the center-left Social Democrats (SPD), the right-of-center, conservative-liberal Free Democrats (FDP), the left-of-center, environmentalist Green party (Greens), the Left party, and, more recently, the far-right "protest" party "Alternative for Germany" (AfD). Moreover, it is an example of a country using a language not as dominant in online discourse as English but still relevant enough to allow for testing of our training data-related arguments, that is, differences in country-level factors and coverage biases affecting the training data. In what can be considered a "next-best" case scenario for LLM public opinion estimation, Germany presents a middle ground between the United States and other societies which are represented in the training data even less, which might pose a challenge for testing synthetic sampling. Findings in LLM public opinion estimation for Germany can be informative for countries with similar characteristics, and even those more underrepresented in the training data in terms of language and society: detecting limitations in LLMs' ability to estimate public opinion in this context would make it likely that this ability is even more limited in more structurally complex, under-researched, or underrepresented contexts.

The social structures dividing the German electorate differ substantially from those characterizing the United States (see, e.g., Lipset & Rokkan, 1967, Brooks et al., 2006, Ford & Jennings, 2020, Sass & Kuhnle, 2023). Moreover, the determinants of voting behavior on the micro-level play out in a different way than in the US context: Partisanship and traditional socio-economic and religious cleavages and their impact on voting behavior have declined (Dalton, 2014, Schmitt-Beck et al., 2022a,b, Berglund et al., 2005, Franklin et al., 2004, Jansen et al., 2013, Elff & Roßteutscher, 2011). At the same time, the socio-cultural dimension (Inglehart, 1977, Schmitt-Beck et al., 2022a) has become more important for voting behavior (Dalton, 2018). As a result of these developments, there are signs of situational issue-voting (Schoen et al., 2017) based on current salient and divisive topics, such as immigration (e.g., Kriesi et al., 2006).

## 3. Data and Methods

### 3.1. Data Collection

*Benchmark data and LLM selection*

In order to examine to what extent LLMs can estimate public opinion in Germany, we simulate a sample of eligible voters in Germany using GPT-3.5. We echo existing research designs in benchmarking the LLM's predicted vote choices against those reported by the survey respondents in the German Longitudinal Election Study (GLES, see Appendix I for details). While surveys are not free from errors, they are currently the best available data source on public opinion on the individual level, allowing us to assess LLM performance for different subgroups of the population.

To ensure comparability with previous studies (Argyle et al., 2023, Bisbee et al., 2024; Dominguez-Olmedo et al., 2023, Hartmann et al., 2023; Motoki et al., 2023, Santurkar et al., 2023), we rely on GPT, which also has the advantage of being one of the largest language models available and being broadly accessible, making it a likely choice for future applications in academia, industry, and by the public. We choose the 2017 German general election because it definitely occurred before the training data cutoff for our specific LLM in June 2021 (OpenAI, n.d., a), with information about the election's context thereby likely included in the training data. If we find limitations in GPT's ability for estimating voting behavior for an election that occurred within the range of its training data, we cannot expect the LLM to perform well in predicting public opinion in contexts beyond its training data.

*Prompt creation*

For the prompts provided to GPT-3.5, we create personas individually simulating each of the 1,905 voting-eligible participants in the 2017 post-election cross-section of the GLES who reported their vote choice (Roßteutscher et al., 2019). The personas include individual-level information on 13 of the most common factors associated with voting behavior as identified in the literature about electoral behavior in Germany (c.f. Schmitt-Beck et al., 2022a, Schmitt-Beck et al., 2022b, Schoen et al., 2017, Klein, 2014). These variables comprise age, gender, educational attainment, income, employment status, residence in East/West Germany, religiosity, ideological left-right self-placement, (strength of) political partisanship, attitude towards immigration, and attitude towards income inequality.[1] Missing values on any of the variables are imputed for n = 377 respondents (20 % of the sample) using multivariate imputation by chained equations (van Buuren & Groothuis-Oudshoorn, 2011). As a robustness check, we adjust the prompt using only the non-imputed variables for the respondents with missing values and compare the results (see Appendix X). We then feed these personas as prompts to GPT-3.5 in German, using the completions-API,[2] alongside the request to complete the last sentence with the respective person's vote choice in the 2017 German parliamentary elections. An example prompt is shown below, translated to English for illustrative purposes (see Appendix II for the German original).

> *I am **28** years old and **female**. I have **a college** degree, a **medium** monthly net household income, and am **working**. I am **not religious**. Ideologically, I am **leaning center-left**. I **rather weakly** identify with **the Green party**. I live in **West** Germany. I think the government should **facilitate** immigration and **take measures to reduce** income disparities.*
> *Did I vote in the 2017 German parliamentary elections and if so, which party did I vote for? I [INSERT]*

*Figure 1: Example prompt (translated from German, variables bolded for emphasis).*

We choose to prompt the LLM in German because the aim of our study is to examine the usability of LLM-generated synthetic samples for public opinion estimation in a non-US- and/or -English context, in order to inform applications outside of the US. Not all local public opinion items are available in English with a faithful translation and testing of concepts. From a normative point of view, requiring an instrument to be translated to English for LLMs to be usable is questionable, as it risks further marginalizing other languages – also when considering LLMs learn from their interactions with human input. Indeed, it is unclear whether English-language prompting would yield better results due to the larger amount of training data. As we have argued, one could conversely expect an LLM to more closely approximate attitudes in the target population when prompted to access those probabilities it has learned from native language training data, as these may be more likely to represent "authentic" attitudes. However, as native-language training data is unequally distributed across target populations, we expect these approximations to be comparatively worse than for a target population whose native language is English (see also Durmus et al., 2023). We leave a comparison of results when using English versus native-language prompting to future research, as it would be out of the scope of the current paper.

[1] For details on the variables in the prompt, see Appendix I and II. For summary statistics, see Appendix VI.

[2] For details, see Appendix III.

*LLM configuration*

Based on the outputs of a pilot test (see Appendix III for details), we calibrate GPT-3.5's text-davinci-003 to a temperature of 0.9 and a response length of maximum 30 tokens.[3] We choose a high temperature to be in line with similar studies (e.g. Argyle et al., 2023, Bisbee et al., 2024) and to simulate the non-determinism in human responses to survey questions (e.g., Zaller, 1992). We collect our data in July 2023 (main sample) and November 2023 (robustness checks). Since the release of GPT-3.5 and its API, OpenAI has performed several changes to both the language model and its data accessibility, including deprecating the possibility of storing token probabilities via the API, that is, the probability with which a sentence is completed with the selected completion token. However, research suggests that first-token probabilities do not always match completions when prompting an LLM with survey questions, especially for sensitive topics that are more likely to induce a refusal from the LLM (Wang et al., 2024). First-tokens also are more sensitive to the prompt format than text output. These limitations make first-token-probabilities an infeasible evaluation metric. To nevertheless account for the probabilistic nature of GPT’s responses beyond a single text completion, we adopt procedures established in multiple imputation (Rubin, 2018). Specifically, we sample five completions per persona and estimate the variance between these samples.[4] By using multiple completions, we can investigate the range and variability of GPT's outputs. This variance analysis helps us grasp the model's behavior and the reliability of its responses, providing insights into the consistency and robustness of the model's text generation. This way, we account for both human (temperature) and LLM (number of samples) randomness in our estimates. Our data thus includes 9525 LLM-generated completions.

[3] See Appendix III for detailed explanations.

*Vote choice extraction*

We then extract the party names from the LLM completions as defined by a set of accepted keywords per party (see Appendix IV), also considering non-voters and invalid votes. 1,427 completions initially did not contain a vote choice. For these, we re-prompt the LLM up to two times, replacing the respective initial completion, resulting in 87 or 0.9% of final completions not containing a vote choice (see Appendix V for details and Appendix X for an investigation of systematic patterns in these personas/completions).

[4] The primary purpose here is to explore the variance in GPT's responses, not to derive formal standard errors, as the assumptions for Rubin's rules, for example, are not fulfilled.

### 3.2. Analysis

We compare the survey-reported and LLM-generated vote choices to investigate the extent to which the responses differ in terms of vote choice as well as how the two data sources weigh the prompt variables in estimating vote choice. This approach allows us to not only assess whether GPT-3.5 is able to estimate the voting behavior of the German general population on aggregate, but also whether it can make equally accurate estimates for different population subgroups.

To tackle our first research question, we compare the aggregate distribution of vote shares across parties according to GPT-3.5 to that based on GLES data. We also estimate multinomial regression models of the prompting variables on voting behavior as reported in GLES and predicted by GPT-3.5, respectively. These models serve two purposes: Relating to our first research question, we evaluate GPT-3.5's predictive performance by comparing its predictions to the predicted values of the GLES-based regression model. We do this by calculating precision, recall, and F1 scores[5] overall and per party, for both the LLM-based predictions and the GLES model predictions. To address our second research question, we also compare the models in terms of effects of specific individual characteristics, as specified by the prompting variables, on voting behavior as reported by GPT-3.5 and GLES, respectively.

For estimating the models, we fit maximum conditional likelihood models based on a neural network with a single hidden layer (Venables & Ripley, 2002). For all regression models, we exclude 78 respondents for whom at least one of the five

[5] Scores range from 0 to 1, with higher values indicating better predictive performance. For a detailed explanation, see Appendix VII.

GPT-samples did not contain an explicit vote choice, in order to ensure comparability across samples,[6] and treat ordinal independent variables with at least five categories as numeric. In order to obtain just one estimate from the five GPT samples, we employ variance estimation as established in multiple imputation research (Rubin 2018). For each analytical method, we calculate each estimate separately for each sample, and then aggregate across the five samples to obtain the average estimate and total standard error. For example, for our regression models, we run five separate regressions, one per sample, and compute the average coefficient and standard error (as $\sqrt{\mu(SE^2_\beta) + 1.2\sigma^2_\beta}$ ) (Rubin 2018) in order to construct confidence intervals.

All analyses are conducted using the software R (R Core Team, 2023), version 4.3.0, especially the packages *tidyverse* (Wickham et al., 2019), *mice* (van Buuren & Groothuis-Oudshoorn, 2011), *rgpt3* (Kleinberg, 2023), *nnet* (Venables & Ripley, 2022), and *marginaleffects* (Arel-Bundock, 2023).

# 4. Results

## 4.1. RQ1: Do LLM-based samples provide similar estimates of voting behavior as national election studies?

*Distribution of vote choice across parties*

On aggregate, the GPT-based distribution of vote shares across parties differs markedly from that of the national election poll. Compared to the GLES sample, GPT-3.5 overestimates the share of Green, Left, and non-voters, while underestimating the share of FDP and AfD voters as well as voters of small parties (see Figure 2). For the two major parties, CDU/CSU and SPD, there are no significant differences.

[6] In Appendix X.3, we show that omitting 73 respondents who GPT classified as non-voters due to a (wrongful) ineligibility from the analysis does not change the results.

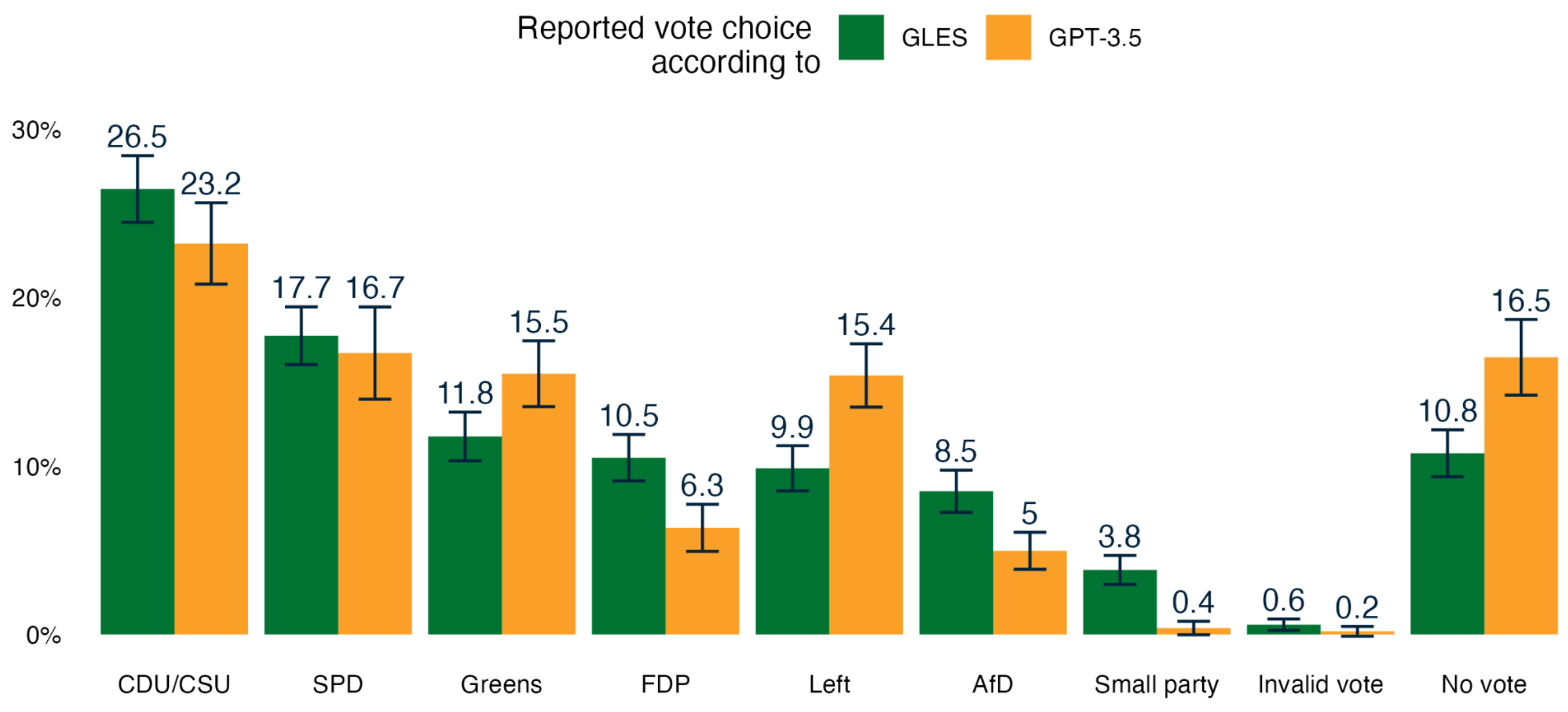

*Figure 2: Distribution of vote shares as estimated by GLES and GPT.*

*Predictive Performance*

Across the five samples drawn from GPT, there is very little variance in terms of whether the GPT prediction matches the vote choice individual respondents reported to GLES. On average, only 46% of GPT-3.5's predictions match the survey data. The F1 scores indicating the LLM's predictive accuracy are best for the CDU/CSU (0.6), followed by SPD and Greens, and much worse for FDP and AfD (around 0.3; see Appendix VII). Comparing these scores to those of predictions based on a multinomial model fitting the prompt variables on *GLES-reported* vote choice, the GLES model creates better predictions than GPT. Given the same demographic and attitudinal information, the GLES model consistently performs better, both overall (0.62) and with regard to specific parties, and most notably for the AfD and FDP (both above 0.5; see

Appendix VII). These differences are informative for both the aggregate and subgroup analyses. For the aggregate, they confirm what the overall distribution of vote shares indicated: The LLM-based estimates of voting behavior are different from survey-based ones, as it is easier for GPT-3.5 to predict voters of Germany's center- and left-leaning parties than right-leaning ones. For subgroups, the higher predictive accuracy of the GLES model justifies benchmarking the predictors of voting behavior according to GPT-3.5 against those in the GLES model.

### 4.2. RQ2: How do LLMs' estimates of voting behavior deviate from national election studies for different subgroups of the population?

Comparing the impact the prompting variables have on actual GPT- versus GLES-reported vote choice in multinomial regressions (see Appendix VIII for the full tables), the models show that GPT-3.5's predictions of vote choice are reliant on certain cues in the prompts, which often do not match the effects the survey data indicates.[7] The model indicates that GPT-3.5 appears to be taking partisanship into particular account when asked to predict people's vote choice. For example, as shown in Figure 3, GPT-3.5 exhibits similar positive effects as the GLES model for SPD, Green, FDP, Left, and AfD partisans on the probability of voting for the respective party. Likewise, it picks up the signal of left-right ideology for the extremes of the party spectrum: the Left and AfD. However, apart from the far-right and -left, GPT-3.5 does not mirror GLES when it comes to the importance of ideology, for example on voting for the CDU/CSU or the Greens. Moreover, when it comes to partisanship, it does not account for *negative* partisanship: For example, the GLES data suggests a systematic underlying pattern between Green and AfD voters – Green partisans are significantly *less* likely to vote for the AfD, and vice versa. In sum, while partisanship and ideology are important factors influencing voting behavior, GPT-3.5 only picks up on broad trends, without regards for more complex dynamics.

[7] It is noticeable that aggregate estimates and individual-level predictive accuracy are better and more similar between GLES and GPT for parties where models of the two data sources share predictors, such as for the CDU/CSU.

Its dominant reliance on party identification as a predictor of vote choice can help explain why GPT-3.5 underestimates the vote shares for FDP and AfD, as observed in section 4.1: Although most partisans indeed vote for the party they identify with (see Table A8 in Appendix IX), only half of the voters of the FDP, AfD, and small parties also identify with their chosen party. Thus, in presence of a partisan cue, GPT-3.5 predicts partisans to vote in line with their party identification. For voters without this cue, its predictions falter.

However, overall, there are more differences than similarities in predictors of vote choice between the LLM-generated and survey data. For the remaining attitudinal variables as well as most demographic indicators, GPT-3.5's predictions assume different mechanisms than what the GLES data suggests, following general patterns identified by previous research on German voting behavior, but not considering the nuances of more complex subgroups. For example, the GPT model, but not the GLES model, suggests residents of East Germany are more likely to vote for the Left or AfD, females are more likely to vote for the Greens or Left, and non-workers less likely to vote for the SPD or FDP (which traditionally have catered to different segments of the working population). Contrary to the GLES model, it does not consider education and income as important factors for distinguishing the likelihood of voting for the Left,

Greens, or FDP versus the AfD, nor the importance of religiosity for distinguishing CDU/CSU from AfD voters.

Similar to what can be observed for (negative) partisanship, GPT-3.5 does not capture the complex effects of attitudes towards inequality and immigration. For example, while the GPT data matches the GLES data in indicating that wanting to limit immigration *decreases* the likelihood of voting for the Greens, the GLES data also indicates that such an attitude *increases* the likelihood of voting for the AfD.

All in all, when considering survey data as ground truth, voting behavior in Germany depends on a different number and kind of factors than GPT-3.5's predictions would suggest. GPT-3.5 bases its predictions on partisanship as well as indicators for common subgroups of voters for a specific party. This finding suggests that GPT-3.5 relies on rather simplified signals in making its predictions, without necessarily considering other, more complex mechanisms in the individual voting-decision making process.

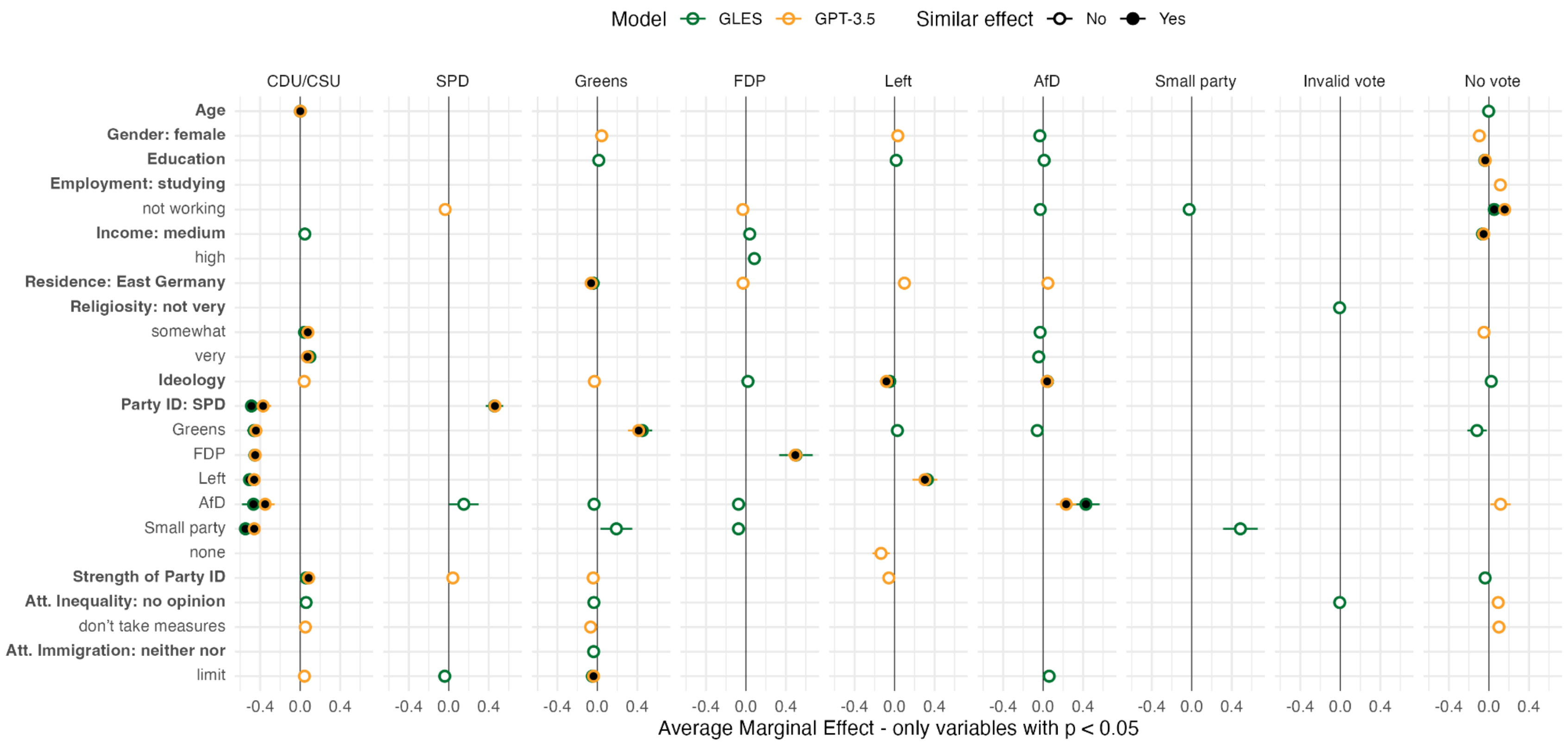

*Figure 3: Average marginal effects of prompt variables on vote choice, as estimated by GLES and GPT.*

*Note: Average marginal effects describe the average of the fitted results of the model after first making individual predictions for each row in the original dataset, mirroring the real data (c.f. Heiss, 2022). "Similar effect" denotes effects that are significant for both the GLES and GPT model and point in the same direction (positive or negative) regardless of magnitude. For reference categories, see Table A7.*

## 5. Discussion

Our study assessed the capabilities of a popular large language model (GPT-3.5 text-davinci-003) in estimating voting behavior for the German federal elections 2017, using the reported vote choices from the respondents of the German Longitudinal Election Study (GLES) data as a benchmark. We created personas simulating every individual respondent in the GLES study. Prompts generated from these personas were then fed to GPT-3.5 via the OpenAI API with a request to complete the personas' vote choice. We compared GPT-3.5's predicted vote choices to respondents' actual vote choices for multiple political parties. Moreover, we conducted a focused subgroup analysis and compared the determinants of voting behavior for the GLES responses with those of the GPT predictions.

Using Germany as an example, we have shown that using LLMs for estimating public opinion in a similar way to surveys cannot simply be generalized beyond the select successful applications in the context of the United States. In our findings, GPT-3.5 overestimated the survey-reported vote shares for the Greens, the Left, and non-voters by a significant margin, while it underestimated the vote shares for FDP and AfD when compared to GLES. The LLM's overall predictive accuracy was modest, with a matching prediction rate of 0.46. It was notably more accurate for voters of the Greens, CDU/CSU, and the Left, but displayed poor predictive power for FDP and AfD voters.

In terms of determining factors that influence voting behavior, GPT-3.5's predictions largely hinged on straightforward indicators such as strong party identification or ideology. However, when compared to the GLES data, it became evident that GPT-3.5 deviated substantially on more complex variables like attitudes towards immigration or economic policy, socio-demographic variables, or the particular dynamics of partisanship. This discrepancy suggests that while GPT-3.5 may capture broad-brush trends tied to partisanship, it tends to miss out on the nuanced, multifaceted factors that sway individual voter choices, thereby limiting its predictive accuracy. As a consequence, relying on LLM-based estimates does not help researchers when predicting voting behavior: Partisans are typically easy to predict as long as they vote in line with their party identification. However, if information on partisanship is necessary for GPT-3.5 to make a prediction, and it cannot evaluate other, more complex relationships in absence of this information, then not only are LLM-based samples not helpful in predicting how non-partisans, weak partisans, or "inconsistent" partisans – all groups who likely swing an election – will vote. They also risk underestimating vote shares for parties with fewer (reported) partisans. Moreover, the absence of mirroring *negative* relationships of factors such as partisanship and immigration attitudes could lead to an underestimation of the popularity of certain parties when applying LLM-based sampling to estimate public opinion. In our case, GPT-3.5 modeled decreasing likelihoods of certain individuals voting for the Greens, without a correct indication of who these individuals would be more likely to vote for (in this case, the AfD), which, consequently, got underestimated.

Naturally, predicting voting behavior in a multi-party system is inherently more difficult than in a two-party system. This challenge remains when transferring the task to LLMs, and therefore is likely one of the reasons why we cannot expect LLMs to work similarly well in all contexts. While researchers hope that LLMs can help them uncover patterns and make predictions where traditional methods struggle, our study underscores the limited applicability of LLM-based synthetic samples. This difficulty is compounded by differences in social structures leading to differential issue conflicts, and by limited nuanced, native-language, and target-population-representing Internet text from which LLMs could learn about these complexities.

Considering the types of text data that were used to train GPT may shed light on its predictive limitations. GPT is trained on a large, but mainstream and not necessarily diverse corpus of text data that includes a selection of websites, books, and other publicly available texts (Brown et al., 2020). As a result, the LLM may be predisposed to make predictions based on generalized or commonly represented political beliefs and more typical, well-researched voter groups, hence struggling with accurately predicting the behavior of voters for the AfD and other non-conforming groups.

This finding underscores the limitation of applying GPT to electoral predictions without accounting for the biases and limitations inherent in its training data. It reaffirms that while GPT can provide certain broad insights, it may not be reliable for nuanced, subgroup-specific political predictions. This is in line with previous issues identified with generative artificial intelligence, which in the context of image generation have been found to reproduce and amplify oftentimes harmful stereotypes and biases (Bianchi et al., 2023, Nicoletti & Bass, 2023, Turk, 2023). Ultimately, using LLMs for estimating public opinion risks reinforcing existing biases. Placing our results in the discourse of existing findings, it becomes clear that LLM-based synthetic samples are only useful in very constrained settings. Researchers considering using LLM-synthetic samples to study public opinion should therefore always critically investigate the applicability of this approach in the specific context to which they want to apply it before doing so, especially when there is no detailed information on the respective model's training data and fine-tuning processes. At this point, there is no clear answer as to whether or when LLM-synthetic samples may be useful for public opinion research – both inside and outside the US. Indeed, even studies considering the US come to diverging conclusions (c.f. Argyle et al., 2023 and Chu et al., 2023 vs. Bisbee et al., 2024, Dominguez-Olmedo et al., 2023, and Santurkar et al., 2023). It thus appears that, similar to surveys conducted with non-probability samples, LLM-based synthetic samples *can* get it right sometimes, but not reliably so. Considering that LLM responses do not represent latent attitudes of an existing target population, but a probability distribution of most-likely next words, even the validity of such measurement may be questioned. It may be argued that LLMs can be useful as long as they provide insights which surveys cannot do. However, as our as well as other studies have shown, they do not meet this expectation so far.

### 5.1. Limitations

Our study encountered several limitations that should be considered when interpreting the results and offer avenues for improvement in future research. First, our study did not test the effect of variable ordering in the prompt on the predictions, which was beyond the scope of our study but could have potentially affected the results (Bisbee et al., 2024). Future research could engage with work on prompt engineering to optimize GPT's predictions. For example, future studies should specify the reference year for time-sensitive variables if it differs from the prompting year (such as age when applied to voting behavior), as Bisbee et al. (2024) suggest. Additionally, while our selection of prompt variables was rooted in existing research on voting behavior in Germany, we acknowledge that other factors might contribute to the voting decision-making process and thus could enhance the predictive power of the LLM. Furthermore, it is unclear how GPT draws on training data in another language than the prompt and completion language. Second, recording token probabilities for the vote choice estimation through the OpenAI API (Argyle et al., 2023) was no longer possible at the time of data collection. This constraint highlights the dependency on the functionalities that API providers offer. Third, the study used text-davinci-003 for its analyses, which may be less efficient and precise than newer language models. However, this choice was made to ensure comparability with existing studies and due to the API availability. The constant “under the hood” changes to and rapid advancement of these language models and their APIs, with the text-davinci-003 model expected to be deprecated by January 2024 (OpenAI, 2023c), raises concerns about the replicability of research such as ours and challenges social scientists to continuously re-evaluate previous findings. Beyond the choice of LLM, results may vary depending on the benchmark survey or specific outcome measured. Investigating the influence of these factors falls outside the scope of this study and is recommended for future research. Fourth, we recognize that our findings for Germany can at best be generalized to Western European socio-political contexts. We argued that our selected case presents a reasonable middle ground for assessing the suitability of LLMs for public opinion research, as it is distinguishable from the United States on the factors we identified as potentially limiting this suitability. While the limitations in LLM public opinion estimation we have found in the German context can be considered unpromising for more structurally complex, under-researched, or disadvantaged societies, such research should be explicitly conducted. However, benchmarking the LLM's responses against reliable individual-level public opinion survey data implies that experiments such as ours can only soundly be conducted in countries that already have a good survey infrastructure, instead of those countries where LLMs could otherwise alleviate a lack of survey research resources. On the other hand, while we treated survey data as ground truth, it is possible that LLMs' predictions better mirror the target population because they do not suffer from the coverage and measurement error of a particular survey. However, we reiterate that in this paper, we were not primarily interested in whether survey- or LLM-generated data is better at accurately predicting actual election outcomes. Both data sources have idiosyncratic error sources leading to differences between their estimates and the actual election results. Comparing errors across data sources would have presented an additional research question that would have been beyond the scope of this paper, but provides an opportunity for future work.

### 5.2. Outlook

This paper contributes to the rapidly growing field of computational social science using LLMs. Many other aspects and conditions of using LLMs for public opinion research are yet to be explored. For example, investigating the suitability of using LLMs for estimating specific minoritized subgroups' attitudes would be a helpful contribution to public opinion research. While this would be possible with the general-population data at hand to a limited extent, benchmarking against special population surveys instead would add value. Moreover, most existing studies have tested whether LLMs are able to "predict the past", benchmarking against survey data from a time included in LLMs' training data. Future research should tackle the question whether LLMs can predict future voting behavior based on past training data, for example by using pre-election panel survey data ahead of an upcoming election for the LLM input and comparing the LLM output to the post-election survey data after the election took place.

Extending the scope to other linguistic, socio-structural, and political contexts, comparative studies could employ cross-national individual-level benchmark datasets. Beyond further examining the contexts in which LLMs can(not) be used for public opinion estimation, such studies could systematically uncover which country-level factors drive this feasibility through multi-level or meta-analyses.

Finally, researchers could explore designing an LLM that is optimized for the purpose of survey research, drawing on comparative evaluations of existing LLMs' performance and the unique requirements of survey research.

## 6. Conclusion

We have shown that in its current state, GPT-3.5 is not suitable for estimating public opinion across (sub)populations, as it exhibits algorithmic bias on two levels. From a cross-sectional perspective, the LLM was unable to pick up on nuances of voter groups, thereby being biased against population subgroups not conforming to the mainstream. From a cross-national perspective, GPT-3.5's performance in estimating voting behavior was not as good for Germany as (some) comparable studies found for the United States. It is likely that predictive performance is even worse for countries, contexts, and populations who are reflected in the LLM training and fine-tuning process even less. The successful application of large language models to public opinion estimation thus is limited to (sub)populations to which their training data is biased – whether this is due to contextual complexity or a lack of linguistic or digital representativity of other populations. More research is necessary to understand what exactly this bias in public opinion estimation depends on and how its sources interact. In sum, GPT-3.5 is better at estimating groups that dominate research and internet data – groups which researchers already know more about, only making LLM-based synthetic samples useful in very limited settings. Researchers need to be aware of these limitations when trying to apply large language models in their work and take care not to reinforce existing biases. Only if large language models are equitable, just, and reflect the population's diversity in a fair manner may we be able to leverage them for estimating public opinion.

## Acknowledgements

The authors would like to thank Max M. Lang, B.Sc., for his assistance in the automated processing of the LLM-generated data, and the FK$^2$RG research group for helpful feedback on earlier drafts of this paper.

Supplementary Material:
Vox Populi, Vox AI? Using Language Models to Estimate German Public Opinion

## Appendices

### I. GLES Questionnaire and GPT Prompt Values

The GLES is based on a multi-stage, stratified, random sample drawn from population registers in Germany (GLES n.d.). Survey participants are interviewed in computer-assisted personal interviews (CAPI).
For details about the primary dataset, see Roßteutscher et al. (2019).

| Variable | GLES Questionnaire (German) | GLES Questionnaire (Translation by GLES) | GPT Prompt Values (German, as prompted) [Translation by authors] |
|---|---|---|---|
| Age | Würden Sie mir bitte sagen, in welchem Jahr Sie geboren wurden? | What year were you born in? | 2017 - year of birth |
| Gender | Intervieweranweisung: Ist die Zielperson männlich oder weiblich?<br><br>(1) Männlich<br>(2) Weiblich | Interviewer instruction: Is the respondent male or female?<br><br>(1) Male<br>(2) Female | **männlich** [male]<br>if Gender = 1<br><br>**weiblich** [female]<br>if Gender = 2 |
| Education | *Schulabschluss*<br>Welchen höchsten allgemeinbildenden Schulabschluss haben Sie?<br><br>(1) Schule beendet ohne Abschluss<br>(2) Hauptschulabschluss, Volksschulabschluss, Abschluss der polytechnischen Oberschule 8. oder 9. Klasse | *School leaving certificate*<br>What's your highest level of general education?<br><br>(1) Finished school without school leaving certificate<br>(2) Lowest formal qualification of Germany's tripartite secondary school system, after 8 or 9 years of schooling | **keinen Schulabschluss**<br>[no degree]<br>if School leaving certificate = 1 \| 9<br><br>**einen Hauptschulabschluss**<br>[Hauptschule degree]<br>if School leaving certificate = 2 |
| | (3) Realschulabschluss, Mittlere Reife, Fachschulreife oder Abschluss der polytechnischen Oberschule 10. Klasse<br>(4) Fachhochschulreife (Abschluss einer Fachoberschule etc.)<br>(5) Abitur bzw. erweiterte Oberschule mit Abschluss 12. Klasse (Hochschulreife)<br>(6) Anderen Schulabschluss, und zwar:<br>(9) Bin noch Schüler<br><br>*Berufliche Bildung*<br>Und welchen beruflichen Ausbildungsabschluss haben Sie? Nennen Sie mir bitte den Kennbuchstaben für den auf Sie zutreffenden Ausbildungsabschluss.<br><br>(A) D - Beruflich-betriebliche Anlernzeit mit Abschlusszeugnis, aber keine Lehre<br>(B) G - Teilfacharbeiterabschluss<br>(C) I - Abgeschlossene gewerbliche oder landwirtschaftliche Lehre<br>(D) B - Abgeschlossene kaufmännische Lehre<br>(E) E - Berufliches Praktikum, Volontariat<br>(F) M - Berufsfachschulabschluss<br>(G) A - Fachakademie-/ Berufsakademieabschluss<br>(H) P - Fachschulabschluss<br>(I) H - Meister, Technikerabschluss<br>(J) K - Fachhochschulabschluss<br>(K) N - Hochschulabschluss: Bachelor<br>(L) O - Hochschulabschluss: Master, Diplom, Magister, Staatsexamen | (3) Intermediary secondary qualification, after 10 years of schooling<br>(4) Certificate fulfilling entrance requirements to study at a polytechnical college<br>(5) Higher qualification, entitling holders to study at a university<br>(6) Other school leaving certificate, please enter:<br>(9) Still at school<br><br>*Vocational and professional training*<br>And what kind of vocational training did you complete? Please name the appropriate letter which corresponds with your vocational training.<br><br>(A) D - On-the-job vocational training with final certificate, but not within a traineeship or apprenticeship scheme<br>(B) G - Compact vocational training course<br>(C) I - Completed trades/crafts or agricultural traineeship<br>(D) B - Completed commercial traineeship<br>(E) E - Work placement/internship<br>(F) M - Specialized vocational college certificate<br>(G) A - Vocational academy certificate<br>(H) P - Technical or vocational college certificate ("Fachschulabschluss")<br>(I) H - Master (craftsman), technician or equivalent college certificate | **einen Realschulabschluss** [Realschule degree]<br>if School leaving certificate = 3 \| 6<br><br>**Abitur** [Abitur degree]<br>if School leaving certificate = 4 \| 5<br><br>**einen Hochschulabschluss** [College degree]<br>if Vocational and professional training = J \| K \| L \| M<br><br>[letter K, N, O, L] |
| | (M) L - Promotion<br>(N) C - Anderen Beruflichen Ausbildungsabschluss, und zwar:<br>(O) F - Noch in beruflicher Ausbildung<br>(P) J - Keine abgeschlossene Ausbildung | (J) K - Polytechnic degree<br>(K) N - University degree, Bachelor<br>(L) O - University degree, Master<br>(M) L - Doctoral degree<br>(N) C - Other vocational training certificate, please enter:<br>(O) F - Still training/studying<br>(P) J - No completed vocational training | |
| Net household income | Wie hoch ist das monatliche Netto-Einkommen IHRES HAUSHALTES INSGESAMT? Ich meine dabei die Summe, die nach Abzug von Steuern und Sozialversicherungsbeiträgen übrig bleibt. Bitte ordnen Sie Ihr Haushaltseinkommen in die Kategorien der Liste ein und nennen Sie mir den Buchstaben.<br><br>(1) B - unter 500 Euro<br>(2) T - 500 bis unter 750 Euro<br>(3) P - 750 bis unter 1000 Euro<br>(4) F - 1000 bis unter 1250 Euro<br>(5) E - 1250 bis unter 1500 Euro<br>(6) H - 1500 bis unter 2000 Euro<br>(7) L - 2000 bis unter 2500 Euro<br>(8) N - 2500 bis unter 3000 Euro<br>(9) R - 3000 bis unter 4000 Euro<br>(10) M - 4000 bis unter 5000 Euro<br>(11) S - 5000 bis unter 7500 Euro<br>(12) A - 7500 bis unter 10000 Euro<br>(13) D - 10000 Euro und mehr | Taken all together, would you please indicate what the monthly net income of your household is? By net income, I mean the amount that you have left after taxes and social security. Please select the monthly net income of your household from one of these groups and tell me the group letter.<br><br>(1) B - Less than 500 euros<br>(2) T - 500 to less than 750 euros<br>(3) P - 750 to less than 1000 euros<br>(4) F - 1000 to less than 1250 euros<br>(5) E - 1250 to less than 1500 euros<br>(6) H - 1500 to less than 2000 euros<br>(7) L - 2000 to less than 2500 euros<br>(8) N - 2500 to less than 3000 euros<br>(9) R - 3000 to less than 4000 euros<br>(10) M - 4000 to less than 5000 euros<br>(11) S - 5000 to less than 7500 euros<br>(12) A - 7500 to less than 10000 euros<br>(13) D - 10000 euros or more | **niedriges** [low]<br>if Net household income = 1 \| 2 \| 3 \| 4 \| 5<br><br>**mittleres** [medium]<br>if Net household income = 6 \| 7 \| 8 \| 9 \| 10<br><br>**hohes** [high]<br>if Net household income = 11 \| 12 \| 13 |
| Employment status | Nun weiter mit der Erwerbstätigkeit und Ihrem Beruf. Was von dieser Liste trifft auf Sie zu?<br><br>(1) Vollzeit berufstätig (mehr als 30 Stunden/Woche)<br>(2) Teilzeit berufstätig (bis 30 Stunden/Woche)<br>(3) Lehrling/Azubi<br>(4) Schüler<br>(5) Student<br>(6) In Umschulung<br>(7) Zurzeit arbeitslos<br>(8) Zurzeit in Kurzarbeit<br>(9) Bundesfreiwilligendienst, Freiwilliges Soziales Jahr (FSJ), Freiwilliges Ökologisches Jahr (FÖJ)<br>(10) Pensionär/Rentner (früher voll berufstätig)<br>(11) In Mutterschutz/Elternzeit<br>(12) Nicht berufstätig (Hausfrau/Hausmann) | Do you currently work in a full-time or part-time job? Which of the descriptions in this list describes your status?<br><br>(1) In full-time employment (more than 30 h/week)<br>(2) In part-time employment (up to 30 h/week)<br>(3) In a traineeship or apprenticeship scheme<br>(4) School student<br>(5) Studying at a polytechnic or university<br>(6) Currently on a retraining course<br>(7) Currently unemployed<br>(8) Currently on short-time working<br>(9) Community service (Bundesfreiwilligendienst, Freiwilliges Soziales Jahr (FSJ), Freiwilliges Ökologisches Jahr (FÖJ))<br>(10) Retirement, on a pension (formerly employed)<br>(11) On maternity leave, parental leave<br>(12) Not in full or part-time employment (Housewife/Househusband) | **nicht berufstätig**<br>[not working]<br>if Employment status = 7 \| 10 \| 12<br><br>**in Ausbildung**<br>[studying/training]<br>if Employment status = 3 \| 4 \| 5 \| 6 \| 9<br><br>**berufstätig** [working]<br>if Employment status = 1 \| 2 \| 8 \| 11 |
| Religiosity | Was würden Sie von sich sagen? Sind Sie überhaupt nicht religiös, nicht sehr religiös, etwas religiös oder sehr religiös?<br><br>(1) Überhaupt nicht religiös<br>(2) Nicht sehr religiös<br>(3) Etwas religiös<br>(4) Sehr religiös | What would you say about yourself, are you not religious at all, not very religious, somewhat religious or very religious?<br><br>(1) Not religious at all<br>(2) Not very religious<br>(3) Somewhat religious<br>(4) Very religious | **überhaupt nicht religiös** [not at all religious] if Religiosity = 1<br><br>**nicht sehr religiös** [not very religious] if Religiosity = 2<br><br>**etwas religiös** [somewhat religious] if Religiosity = 3<br><br>**sehr religiös** [very religious] if Religiosity = 4 |
| Left-right self-placement | Und wie ist das mit Ihnen selbst? Wo würden Sie sich auf der Skala von 1 bis 11 einordnen?<br>(1) 1 links<br>(2) 2<br>(3) 3<br>(4) 4<br>(5) 5<br>(6) 6<br>(7) 7<br>(8) 8<br>(9) 9<br>(10) 10<br>(11) 11 rechts | Where would you place yourself on this scale?<br>(1) 1 Left<br>(2) 2<br>(3) 3<br>(4) 4<br>(5) 5<br>(6) 6<br>(7) 7<br>(8) 8<br>(9) 9<br>(10) 10<br>(11) 11 Right | **stark links** [strongly left] if Left-right self-placement = 1 \| 2<br><br>**mittig links** [center-left] if Left-right self-placement = 3 \| 4<br><br>**in der Mitte** [in the middle] if Left-right self-placement = 5 \| 6 \| 7<br><br>**mittig rechts** [center-right] if Left-right self-placement = 8 \| 9<br><br>**stark rechts** [strongly right] |
| | | | if Left-right self-placement = 10 \| 11 |
| Party identification | Und nun noch einmal kurz zu den politischen Parteien. In Deutschland neigen viele Leute längere Zeit einer bestimmten politischen Partei zu, obwohl sie auch ab und zu eine andere Partei wählen. Wie ist das bei Ihnen: Neigen Sie - ganz allgemein gesprochen - einer bestimmten Partei zu? Und wenn ja, welcher?<br><br>[Liste für Interviewer]<br><br>(1) CDU/CSU<br>(2) CDU<br>(3) CSU<br>(4) SPD<br>(5) FDP<br>(6) GRÜNE<br>(7) DIE LINKE<br>(322) AfD<br>(801) andere Partei, und zwar<br>(808) keine Partei | Now, let's look at the political parties. In Germany, many people lean towards a particular party for a long time, although they may occasionally vote for a different party. How about you, do you in general lean towards a particular party? If so, which one?<br><br>[List for interviewer]<br><br>(1) CDU/CSU<br>(2) CDU<br>(3) CSU<br>(4) SPD<br>(5) FDP<br>(6) GRÜNE<br>(7) DIE LINKE<br>(322) AfD<br>(801) [other party: specify]<br>(808) [no party] | **mit der Partei CDU/CSU** [CDU/CSU]<br>if Party identification = 1 \| 2 \| 3<br><br>**mit der Partei SPD** [SPD]<br>if Party identification = 4<br><br>**mit der Partei FDP** [FDP]<br>if Party identification = 5<br><br>**mit der Partei Bündnis 90/Die Grünen** [Greens]<br>if Party identification = 6<br><br>**mit der Partei Die Linke** [Left]<br>if Party identification = 7<br><br>**mit der Partei AfD** [AfD]<br>if Party identification = 322<br><br>**mit einer Kleinpartei** [small/other party]<br>if Party identification = 801<br><br>**mit keiner Partei** [not with any party]<br>if Party identification = 808 |
| | | | |
| Strength of party identification | Wie stark oder wie schwach neigen Sie - alles zusammengenommen - dieser Partei zu: sehr stark, ziemlich stark, mäßig, ziemlich schwach oder sehr schwach?<br><br>(1) Sehr stark<br>(2) Ziemlich stark<br>(3) Mäßig<br>(4) Ziemlich schwach<br>(5) Sehr schwach | All in all, how strongly or weakly do you lean toward this party: very strongly, fairly strongly, moderately, fairly weakly or very weakly?<br><br>(1) Very strongly<br>(2) Fairly strongly<br>(3) Moderately<br>(4) Fairly weakly<br>(5) Very weakly | **sehr schwach** [very weakly] if Strength of party identification = 5<br><br>**ziemlich schwach** [rather weakly] if Strength of party identification = 4<br><br>**mäßig** [moderately] if Strength of party identification = 3<br><br>**ziemlich stark** [rather strongly] if Strength of party identification = 2<br><br>**sehr stark** [very strongly] if Strength of party identification = 1 |
| East/West Germany | (0) Ostdeutschland<br>(1) Westdeutschland | [coded by interviewer]<br>0 East Germany<br>1 West Germany | **Westdeutschland** [West Germany] 0<br>**Ostdeutschland** [East Germany] 1 |
| | | | |
| Immigration | Und wie ist Ihre Position zum Thema Zuzugsmöglichkeiten für Ausländer? Bitte benutzen Sie diese Skala.<br>(1) 1 Zuzugsmöglichkeiten für Ausländer sollten erleichtert werden<br>(2) 2<br>(3) 3<br>(4) 4<br>(5) 5<br>(6) 6<br>(7) 7<br>(8) 8<br>(9) 9<br>(10) 10<br>(11) 11 Zuzugsmöglichkeiten für Ausländer sollten eingeschränkt werden | And what position do you take on immigration for foreigners? Please use the scale.<br>(1) 1 Immigration for foreigners should be easier<br>(2) 2<br>(3) 3<br>(4) 4<br>(5) 5<br>(6) 6<br>(7) 7<br>(8) 8<br>(9) 9<br>(10) 10<br>(11) 11 Immigration for foreigners should be more difficult | **einschränken** [limit]<br>if Immigration = 7 \| 8 \| 9 \| 10 \| 11<br><br>**weder erleichtern noch einschränken** [neither nor]<br>if Immigration = 6<br><br>**erleichtern** [facilitate]<br>if Immigration = 1 \| 2 \| 3 \| 4 \| 5 |
| Inequality | Es gibt zu verschiedenen politischen Themen unterschiedliche Meinungen. Wie ist das bei Ihnen: Was halten Sie von folgenden Aussagen? Bitte antworten Sie anhand der Liste.<br><br>(D) Die Regierung sollte Maßnahmen ergreifen, um die Einkommensunterschiede zu verringern.<br><br>(1) Stimme voll und ganz zu<br>(2) Stimme eher zu<br>(3) Teils/teils<br>(4) Stimme eher nicht zu | There are various opinions on different political issues. What do you think of the following statements?<br>Please use the list.<br><br>(D) The government should take measures to reduce the differences in income levels.<br><br>(1) Strongly agree<br>(2) Agree<br>(3) Neither agree nor disagree<br>(4) Disagree<br>(5) Strongly disagree | **keine Maßnahmen ergreifen** [don't take measures]<br>if Inequality= 4 \| 5<br><br>**habe keine Meinung dazu, ob die Regierung Maßnahmen ergreifen sollte** [no opinion]<br>if Inequality= 3<br><br>**Maßnahmen ergreifen** [take measures]<br>if Inequality= 1 \| 2 |
| | (5) Stimme überhaupt nicht zu | | |

*Table A1: GLES variables and corresponding prompt variables.*

## II. Example Prompts

### 1. Main sample

| English (translation) | German (as prompted) |
| --- | --- |
| I am **28** years old and **female**. I have a **college** degree, a **medium** monthly net household income, and am **working**. I am **not religious**. Ideologically, I am **leaning center-left**. I **rather weakly** identify with **the Green party**. I live in **West** Germany. I think the government should **facilitate** immigration and **take measures to reduce** income disparities.<br>Did I vote in the 2017 German parliamentary elections and if so, which party did I vote for? I [INSERT] | Ich bin **28** Jahre alt und weiblich. Ich habe **einen Hochschulabschluss**, ein **mittleres** monatliches Haushalts-Nettoeinkommen und bin **berufstätig**. Ich bin **nicht religiös**. Politisch-ideologisch ordne ich mich **mittig links** ein. Ich identifiziere mich **ziemlich schwach** mit der Partei **Bündnis 90/Die Grünen**. Ich lebe in **Westdeutschland**. Ich finde, die Regierung sollte die Einwanderung **erleichtern** und **Maßnahmen ergreifen**, um die Einkommensunterschiede zu verringern.<br>Habe ich bei der Bundestagswahl 2017 gewählt und wenn ja, welcher Partei habe ich meine Zweitstimme gegeben?<br>Ich habe [INSERT] |

*Table A2a: Example prompt for the main study and English translation (variables bolded).*

*Note: We decided not to include “gewählt” (voted) as a suffix in the prompt, using the [MASK] instead of [INSERT] request, as it might bias the output against non-voters by reducing the likelihood of GPT completing the sentence with “nicht” (not) or “ungültig” (invalid) due to German semantics. We leave the further exploration of these effects to prompt engineering researchers.*

2. **Robustness sample (respondents with missing values)**

Any sentence except the first one was omitted if the respective variable was missing for the respondent.

| English (translation) | German (as prompted) |
|---|---|
| I am **28** years old. I am **female**. I have a **college** degree. I have a **medium** monthly net household income. I am **working**. I am **not religious**. Ideologically, I am **leaning center-left**. I **rather weakly** identify with **the Green party**. I live in **West** Germany. I think the government should **facilitate** immigration. I think the government should **take measures to reduce** income disparities.<br>Did I vote in the 2017 German parliamentary elections and if so, which party did I vote for? I [INSERT] | Ich bin **28** Jahre alt. Ich bin weiblich. Ich habe **einen Hochschulabschluss**. Ich habe ein **mittleres** monatliches Haushalts-Nettoeinkommen. Ich bin **berufstätig**. Ich bin **nicht religiös**. Politisch-ideologisch ordne ich mich **mittig links** ein. Ich identifiziere mich **ziemlich schwach** mit der Partei **Bündnis 90/Die Grünen**. Ich lebe in **Westdeutschland**. Ich finde, die Regierung sollte die Einwanderung **erleichtern**. Ich finde, die Regierung sollte **Maßnahmen ergreifen**, um die Einkommensunterschiede zu verringern.<br>Habe ich bei der Bundestagswahl 2017 gewählt und wenn ja, welcher Partei habe ich meine Zweitstimme gegeben?<br>Ich habe [INSERT] |

*Table A2b: Example prompt for robustness checks and English translation (variables bolded).*

## III. Model Configuration Parameters & Pilot

According to OpenAI, the difference between the chat completions-API (the API accessing the model embedded in ChatGPT and therefore commonly referred to as “ChatGPT”) and the completions-API lies in the underlying models and cost, with the chat completions-API offering access to the more capable GPT-4 and cost-effective GPT-3.5-turbo, which corresponds to the performance of the completion-API’s text-davinci-003, which we used, at a fraction of the cost (OpenAI, n.d., b). Text-davinci-003 is a version of GPT-3.5 optimized for efficient text completion (OpenAI, n.d., a).

We opted for specifying the randomness (temperature) rather than restricting the completion-sample to the tokens above a certain probability threshold (parameter top-p), as recommended by OpenAI. We further opted not to specify a penalty for the prompt information provided on party identification and/or ideology, leaving the details of this for further research in prompt engineering. The following configurations were tested sequentially on a subsample of five personas (see supplementary material for output data):

1. **Max. Tokens**

Specifies the maximum amount of tokens a completion may contain. One token corresponds to about four letters in the English language.
Default values for other parameters: temperature = 1, n=1

- Test 1: maxtoken = 20: 1/3 did not contain party; 20 tokens used per completion
- Test 2: maxtoken = 30: works as desired, 1 "insert" but contains party; 10-25 tokens used per completion
- Test 3: maxtoken = 40: one contradictory and one ambiguous case, but complete information; 20-30 tokens used per completion
- Test 4: maxtoken = 50: complete information; 30-40 tokens used per completion

30 tokens provide ample space for GPT to complete the prompt with a full sentence containing vote choice. This does not imply that all of GPT’s completions will be 30 tokens in length. Indeed, our analyses reveal that a substantial portion of completions break off despite using less than 30 tokens.

**→ DECISION ON TOKENS: 30**

2. **Temperature**

Specifies the randomness of possible completions by specifying the sampling strategy from the underlying distribution. 0 corresponds to no randomness (deterministic), with repetitive completions, 1 corresponds to complete randomness with maximum variation. Default in OpenAI Playground: 1. For detailed explanations, see Argyle et al. (2023).
Values for other parameters: maxtokens = 30, n=1

- Test 5: temperature = 0.9 (rgpt3 package default): complete information, 1/3 incomplete sentence, 25-30 tokens used per completion
- Test 6: temperature = 0.7 (Argyle et al. 2023): complete information, complete sentences, 25-30 tokens used per completion. Differences in person with ambiguous predictors

**→ DECISION ON TEMPERATURE: 0.9**

3. **Multiple completions vs. Best of**

*N* specifies the number of completions per prompt (default: 1). *Best of* determines the space of possibilities from which to select the completion with the highest probability. Generates best_of completions server-side and returns the "best" (the one with the highest log probability per token).
Values for other parameters: maxtokens = 30, temperature = 0.9

- Test 7: n = 5:
  - Records 5 completions
  - Returns warning: To avoid an `invalid_request_error`, `best_of` was set to equal `n`
- Test 8: best_of = 5 Records only 1 completion

**→ DECISION ON N vs BEST OF: N = 5, best_of = 1 to account for inability to store token probabilities.**
Function will force value to default to n.

## IV. Accepted completions for party matching

Small parties that were listed on the ballot for the 2017 election but were not represented in the 18th German Bundestag (2013-2017) are summarized as “Small party”.

| **Party / GLES reported vote** **(**Question q19ba;<br>[brackets]: translation) | **GPT completion contains**<br>(case-insensitive;<br>*asterisk*: embedded within any word;<br>**bold**: conceptually equivalent wordings) |
|---|---|
| CDU/CSU | CDU, CSU, CDU/CSU, Union, *christ* |
| SPD | SPD, *sozialdemokrat* |
| Bündnis 90/Die Grünen [Greens] | *Grün*, 90, Bündnis |
| FDP | FDP, freie, *liberal* |
| Die Linke [Left] | *link* [confirmed by manual check] |
| AfD | AfD, Alternative [confirmed by manual check] |
| Andere Partei [other / small party] | Andere [confirmed by manual check]<br>**Kleinpartei** [confirmed by manual check]<br>***any small party names, e.g. Piraten*** [confirmed by manual check] |
| Ungültig gewählt [invalid vote] | [confirmed by manual check]<br>ungültig<br>**keine Zweitstimme** |
| Nicht gewählt [did not vote] | [confirmed by manual check]<br>nicht, **keine Partei, weder gewählt noch eine Zweitstimme abgegeben** |

*Table A3: Keywords used for matching between GPT completions and German political parties / GLES vote choice.*

## V. Documentation of manual checks

- If the completion contained "a left party" without specification for "Die Linke", it was recoded as NA and resampled
- Manual checks were performed if multiple party names were extracted and/or "Erststimme" (primary vote) was mentioned, in order to extract the correct party
  - If multiple parties were named and no distinction between Erst- and Zweitstimme was made, the second-named party was assumed to be the Zweitstimme
  - If "Erststimme" and "Zweitstimme" were mentioned and it was not clear which vote the party name referred to, it was recoded as NA
  - If "Erststimme" but no "Zweitstimme" was mentioned, it was recoded as NA
  - If the completion contained "meine Stimme" (my vote) and only one party without an indication of whether this was Erst- or Zweitstimme, Zweitstimme was assumed and the vote recorded accordingly
- Manual checks were performed for all completions that couldn't be matched with any vote choice automatically
- Hallucinations were recoded as NAs and resampled. Reasoning: face-validity (no party on the ballot in Germany or perfectly matchable to one party)
- If the completion contained the equivalent of "voted" followed by "did not vote", it was recoded as "did not vote", as it became evident that completing the sentence "Ich habe" with "gewählt" was a common/meaningless default before elaborating on the choice made.
- Notes:
  - 1 completion contained "Parteienalternative" which was thus matched with AfD, but could also have been matched with "Other party"
  - 1 completion contained a party that cannot be further named [breakoff], which was wrongly matched as "did not vote" but could have been an NA

| | **Sample 1** | **Sample 2** | **Sample 3** |
|---|---|---|---|
| **Total completions** | 9525 | 1427 | 281 |
| **Total modified** | 653 (6.9%) | 107 (7.5%) | 27 (9.6%) |
| **NAs (after modification)** | 1427 (14.9%) | 281 (19.7%) | 89 (31.7%) |
| **Robustness: Total completions** | 1885 | 461 | 235 |
| **Robustness: Total modified** | 78 (4.1%) | 20 (4.3%) | 10 (4.3%) |
| **Robustness: NAs (after modification)** | 461 (24.5%) | 235 (51%) | 141 (60%) |

*Table A4: Completions obtained from GPT, modifications and missing completions in main and robustness samples.*

## VI. Summary Statistics, GLES Data (Main Sample)

| Variable | N | Mean / Prop. | Std. Dev. | Min. | Median | Max. |
|---|---|---|---|---|---|---|
| **Age** | 1905 | 51 | 18 | 18 | 52 | 95 |
| **Gender** | 1905 | | | | | |
| male | 1001 | 53% | | | | |
| female | 904 | 47% | | | | |
| **Education** | 1905 | 2.5 | 1.2 | 0 | 2 | 4 |
| **Employment Status** | 1905 | | | | | |
| not working | 671 | 35% | | | | |
| studying | 143 | 8% | | | | |
| working | 1091 | 57% | | | | |
| **Income** | 1905 | | | | | |
| low | 349 | 18% | | | | |
| medium | 1324 | 70% | | | | |
| high | 232 | 12% | | | | |
| **Residence** | 1905 | | | | | |
| West Germany | 1289 | 68% | | | | |
| East Germany | 616 | 32% | | | | |
| **Religiosity** | 1905 | | | | | |
| not at all | 668 | 35% | | | | |
| not very | 320 | 17% | | | | |
| somewhat | 689 | 36% | | | | |
| very | 228 | 12% | | | | |
| **LR-Ideology** | 1905 | -0.3 | 0.81 | -2 | 0 | 2 |
| **Party ID** | 1905 | | | | | |
| CDU/CSU | 540 | 28% | | | | |
| SPD | 356 | 19% | | | | |
| Greens | 185 | 10% | | | | |
| FDP | 98 | 5% | | | | |
| Left | 159 | 8% | | | | |
| AfD | 83 | 4% | | | | |
| Small party | 23 | 1% | | | | |
| none | 461 | 24% | | | | |
| **Strength of Party ID** | 1905 | 2.7 | 1.7 | 0 | 3 | 5 |
| **Att. Inequality** | 1905 | | | | | |
| act | 1464 | 77% | | | | |
| no opinion | 251 | 13% | | | | |
| don't act | 190 | 10% | | | | |
| **Att. Immigration** | 1905 | | | | | |
| facilitate | 603 | 32% | | | | |
| neither nor | 350 | 18% | | | | |
| limit | 952 | 50% | | | | |
| **Vote Choice (GLES)** | 1905 | | | | | |
| CDU/CSU | 504 | 26% | | | | |
| SPD | 338 | 18% | | | | |
| Greens | 224 | 12% | | | | |
| FDP | 200 | 10% | | | | |
| Left | 188 | 10% | | | | |
| AfD | 162 | 9% | | | | |
| Small party | 73 | 4% | | | | |
| Invalid vote | 11 | 1% | | | | |
| No vote | 205 | 11% | | | | |
| **Imputed** | 1905 | | | | | |
| no | 1528 | 80% | | | | |
| yes | 377 | 20% | | | | |

*Table A5: Summary statistics of prompt input variables, main sample.*

## VII. Model Performance

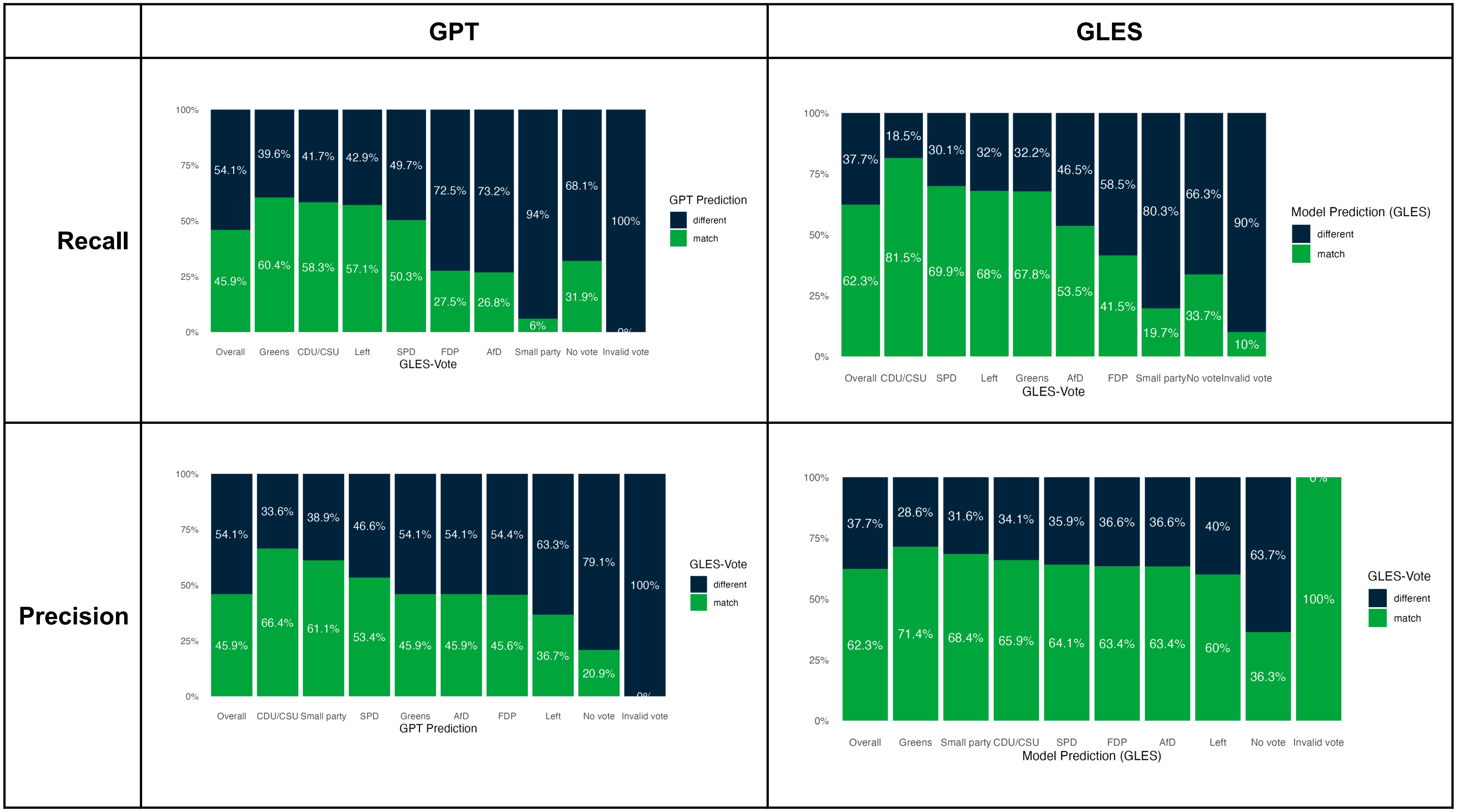

*Figure A1: Model performance of GPT predictions vs. predictions based on multinomial regression of prompt variables on GLES-reported vote choice.*

Recall refers to the number of correct predictions for a given party (true positives) divided by the number of actual votes for that party (true positives + false negatives), measuring the share of votes for a party as reported in the GLES that were correctly identified by GPT/the GLES-based model.

Precision refers to the number of correct predictions for a given party (true positives) divided by the number of all predicted votes for that party (true + false positives), measuring the share of GPT/GLES predicted votes for a given party that were correct.

F1 scores evaluate machine learning model accuracy by considering model precision and recall. The F1 score is defined as 2*precision*recall / (precision + recall), with a range of [0;1].

| | **F1 Score** $\frac{2 \times precision \times recall}{precision + recall}$ | |
|---|---|---|
| **Party** | **GPT** | **GLES Model** |
| *Overall* | *0.46* | *0.62* |
| CDU/CSU | 0.62 | 0.73 |
| SPD | 0.52 | 0.67 |
| Greens | 0.52 | 0.70 |
| Left | 0.45 | 0.64 |
| FDP | 0.34 | 0.50 |
| AfD | 0.33 | 0.58 |
| No vote | 0.25 | 0.35 |
| Small party | 0.11 | 0.31 |
| Invalid | 0 | 0.18 |

*Table A6: Model accuracy (F1 scores) of GPT predictions vs. predictions based on multinomial regression of prompt variables on GLES-reported vote choice.*

## VIII. Multinomial Regression Model

Values for GPT are average values across separate regressions for each of the five samples of completions.

**Effect on Vote Choice – GLES vs. GPT (Reference: CDU/CSU)**

| Independent Variables | | | SPD | Greens | FDP | Left | AfD | Small party | Invalid vote | No vote |
|---|---|---|---|---|---|---|---|---|---|---|
| Age | | GLES | 0.000 | -0.001 | -0.007 | -0.010 | -0.003 | -0.024 | -0.057 * | -0.037 *** |
| | | GPT | -0.016 | -0.027 | -0.014 | -0.027 * | -0.019 | -0.030 | -0.010 | -0.019 |
| Gende (Ref.: male) | female | GLES | -0.100 | -0.102 | 0.023 | -0.475 | -0.833 ** | -0.239 | -1.492 | -0.181 |
| | | GPT | 0.287 | 0.666 * | -0.131 | 0.480 | -0.048 | 0.326 | -0.674 | -0.893 *** |
| Education | | GLES | 0.059 | 0.321 ** | -0.020 | 0.364 ** | 0.156 | 0.106 | 0.253 | -0.417 *** |
| | | GPT | -0.003 | 0.172 | 0.130 | -0.013 | 0.022 | 0.009 | 0.101 | -0.397 ** |
| Employment (Ref.: working) | studying | GLES | -0.152 | 0.099 | -0.324 | -0.445 | -0.953 | -0.167 | 0.829 | -0.177 |
| | | GPT | -0.102 | -0.096 | -0.297 | -0.057 | -1.209 | 0.603 | -0.358 | 1.256 * |
| | not working | GLES | 0.399 | -0.049 | 0.302 | 0.442 | -0.268 | -0.767 | 1.365 | 0.735 * |
| | | GPT | -0.293 | -0.209 | -0.732 | -0.075 | -0.329 | -0.267 | 0.076 | 1.560 *** |
| Income (Ref.: low) | medium | GLES | -0.497 | -0.307 | 0.302 | -0.539 | -0.540 | -0.671 | -0.624 | -1.006 *** |
| | | GPT | -0.234 | -0.233 | -0.061 | -0.441 | -0.359 | -0.756 | -0.614 | -0.824 * |
| | high | GLES | -0.807 | -0.195 | 0.766 | -0.843 | -0.875 | -1.155 | -2.710 | -1.037 * |
| | | GPT | -0.406 | -0.418 | -0.073 | -0.883 | -0.353 | -1.778 | -0.186 | -0.933 |
| Residence (Ref.: West Germany) | East Germany | GLES | -0.331 | -0.843 ** | -0.194 | 0.263 | 0.269 | 0.275 | 0.874 | 0.276 |
| | | GPT | 0.518 | -0.439 | -0.732 | 1.507 *** | 1.677 *** | 0.280 | 0.305 | 0.197 |
| Religiosity (Ref.: not at all) | not very | GLES | -0.201 | 0.179 | 0.276 | 0.067 | -0.668 | 0.117 | -3.483 | -0.447 |
| | | GPT | -0.410 | -0.380 | -0.461 | -0.534 | -0.426 | -0.529 | -0.512 | -0.767 |
| | somewhat | GLES | -0.244 | -0.184 | -0.256 | -0.641 | -0.829 ** | -0.526 | -0.224 | -0.408 |
| | | GPT | -0.883 * | -1.465 *** | -1.122 * | -1.346 *** | -0.377 | -0.664 | -1.323 | -1.436 *** |
| | very | GLES | -1.022 ** | -0.460 | -0.744 * | -1.303 * | -1.460 ** | 0.052 | -2.900 | -0.776 * |
| | | GPT | -1.515 ** | -1.617 ** | -0.905 | -1.553 ** | -0.318 | -1.542 | -1.544 | -1.003 * |
| LR-Ideology | | GLES | -0.437 ** | -0.554 ** | 0.197 | -1.154 *** | 0.766 *** | -0.462 | -0.631 | 0.162 |
| | | GPT | -0.787 ** | -1.331 *** | -0.269 | -1.938 *** | 0.919 ** | -0.680 | -0.287 | -0.677 ** |

cont.

| Independent Variables | | Effect on Vote Choice – GLES vs. GPT (Reference: CDU/CSU) | | | | | | | |
|---|---|---|---|---|---|---|---|---|---|
| | | SPD | Greens | FDP | Left | AfD | Small party | Invalid vote | No vote |
| Party ID (Ref.: CDU/CSU) | SPD | 1.064 | 1.885 | 3.178 | 3.384 | 1.828 | -1.377 | 3.489 | 0.205 |
| | | -7.848 | -7.043 | -3.839 | -7.208 | -6.585 | -2.030 | -1.673 | -3.380 |
| | Greens | 0.003 | 4.195 | 5.674 | -1.611 | -1.826 ** | 2.201 | -1.375 | -9.351 * |
| | | 0.653 | -3.556 | 2.219 | -2.114 | -2.775 | 1.455 | 0.023 | 0.839 |
| | FDP | -0.317 | 0.721 | 5.394 ** | 1.896 | -0.017 | -5.423 | 0.132 | 3.549 |
| | | -4.024 | -4.153 | -2.779 | -0.526 | -5.213 | -0.189 | -0.603 | 0.596 |
| | Left | -2.236 | 7.655 | 4.615 | 1.456 | 1.244 | -0.259 | 1.161 | -0.049 |
| | | -0.344 | -2.915 | -1.054 | -3.084 | 0.683 | 1.790 | 0.201 | 2.835 |
| | AfD | 6.891 | -1.832 * | -1.548 ** | -1.154 | 4.123 | -2.205 * | -2.487 | 2.558 |
| | | -1.040 | -4.932 | -1.126 | -5.672 | -9.735 ** | -0.501 | -0.679 | -2.304 |
| | Small party | -0.877 *** | 8.022 | -0.184 *** | 0.427 | -1.809 | -1.845 | 4.214 | -0.741 |
| | | 3.897 | -5.410 | 0.572 | -3.173 | -2.134 | -0.349 | 0.335 | 1.884 |
| | none | -0.588 | 1.215 | 1.037 | 1.320 | -0.580 | -1.940 | -0.175 | -1.527 |
| | | -2.881 | -5.475 ** | -1.956 | -5.762 *** | -4.704 ** | 0.895 | 1.056 | -0.739 |
| Strength of Party ID | | -0.954 * | -0.309 | -0.179 | -1.167 * | -0.758 * | -1.417 ** | -1.405 | -1.020 *** |
| | | -1.767 ** | -2.564 *** | -1.227 | -2.658 *** | -1.963 *** | -0.504 | -0.762 | -1.439 ** |
| Party ID x Strength (Ref.: CDU/CSU) | SPD | 1.241 * | 0.351 | -0.509 | 0.639 | 0.033 | 1.238 | 0.480 | 0.630 |
| | | 4.291 ** | 3.435 ** | 1.829 | 3.503 * | 2.832 | 1.911 | 1.422 | 2.007 |
| | Greens | 0.936 | 0.308 | -1.448 | 1.893 * | -0.410 | 0.237 | 0.905 | 2.811 * |
| | | 0.898 | 3.018 ** | 0.207 | 1.263 | 0.944 | 0.810 | 0.899 | 0.847 |
| | FDP | 0.561 | 0.384 | -0.423 | 0.795 | 0.353 | 2.251 | 0.444 | -1.155 |
| | | 1.707 | 2.280 | 2.769 | 0.112 | 2.050 | 1.071 | 0.973 | 0.952 |
| | Left | 1.710 | -1.347 | -0.916 | 1.972 * | 0.602 | 1.412 | 1.230 | 1.013 |
| | | 1.980 | 2.729 | 0.274 | 3.488 | 0.950 | 1.467 | 1.592 | 1.154 |
| | AfD | -1.923 | -0.013 | -0.191 | 0.427 | 0.369 | 0.190 | 2.511 | 0.271 |
| | | 0.634 | 1.063 | 0.763 | 2.182 | 4.016 *** | 0.759 | 0.437 | 1.421 |
| | Small party | -0.533 *** | 1.368 | 0.982 *** | 3.328 | 4.656 ** | 6.076 *** | 4.457 ** | 4.779 *** |
| | | -0.963 | 4.514 ** | 1.124 | 3.192 * | 3.456 * | 4.610 ** | 1.499 | 2.532 |

cont.

| Independent Variables | | Effect on Vote Choice – GLES vs. GPT (Reference: CDU/CSU) | | | | | | | |
|---|---|---|---|---|---|---|---|---|---|
| | | SPD | Greens | FDP | Left | AfD | Small party | Invalid vote | No vote |
| Att. Inequality (Ref.: take measures) | no opinion | -0.350 | -1.076 ** | -0.427 | -0.984 * | -0.519 | -0.226 | -3.204 | -0.500 |
| | | -0.692 | -0.730 | 0.051 | -0.315 | 0.182 | -0.623 | -0.411 | 0.739 * |
| | don't take measures | -0.888 * | -0.692 | 0.225 | -0.810 | -0.145 | -0.651 | -0.111 | -0.503 |
| | | -1.893 *** | -2.228 *** | -0.317 | -2.049 ** | -0.105 | -2.664 | -1.973 | -0.065 |
| Att. Immigration (Ref.: facilitate) | neither nor | -0.572 | -0.843 * | 0.173 | -0.456 | 0.438 | -0.286 | 1.856 | 0.088 |
| | | -0.387 | -0.432 | -0.024 | -0.253 | -1.423 | -0.035 | -0.633 | -0.862 |
| | limit | -0.513 * | -0.987 *** | 0.045 | 0.000 | 1.435 *** | -0.219 | 2.014 | 0.211 |
| | | -0.890 ** | -1.220 *** | -0.238 | -0.712 * | -0.155 | -0.176 | -0.823 | -0.739 * |
| Intercept | | 1.224 | -1.689 | -1.175 | -1.671 | 0.250 | 2.753 | -1.528 | 5.117 *** |
| | | 5.392 * | 8.490 *** | 3.080 | 8.353 *** | 4.816 | -3.755 | -2.624 | 6.081 *** |
| N | | 1827 | | | | | | | |
| * | p<0.05 | | | | | | | | |
| ** | p<0.01 | | | | | | | | |
| *** | p<0.001 | | | | | | | | |

Multinomial regression.
Reported values represent beta coefficients.

*Table A7: Results of a multinomial regression of prompt variables on vote choice.*

## IX. Distribution of partisanship and choice in regression samples

FDP and AfD are the parties with the lowest shares of partisans in the survey data, with the reported vote share for these parties being twice as high as their respective share of partisans. Moreover, the share of non-voters is disproportionately high among people identifying with the AfD or a small party, and non-partisans.

| **Party ID**<br><br>**Vote** | **CDU/ CSU** (28.9%) | **SPD** (19%) | **Greens** (10.1%) | **FDP** (5.3%) | **Left** (8.6%) | **AfD** (4.5%) | **Small party** (1.1%) | **No party** (22.4%) |
|---|---|---|---|---|---|---|---|---|
| **CDU/CSU** (26.9%) | **70.8%** | 6% | 5.4% | 11.5% | 1.9% | 2.4% | 0% | 17.1% |
| **SPD** (18%) | 2.8% | **67%** | 8.1% | 2.1% | 6.3% | 1.2% | 0% | 13% |
| **Greens** (11.7%) | 2.3% | 6% | **75.1%** | 3.1% | 5.1% | **0%** | 5% | 7.3% |
| **FDP** (10.3%) | 9.7% | 3.2% | 1.6% | **74%** | 1.3% | 0% | 0% | 12.2% |
| **Left** (9.9%) | 0.2% | 4.9% | 4.3% | 2.1% | **67.7%** | 0% | 0% | 11.2% |
| **AfD** (8.5%) | 5.3% | 2.6% | **0%** | 3.1% | 3.2% | **77.1%** | 5% | 11% |
| **Small party** (3.6%) | 1.1% | 1.7% | 3.2% | 2.1% | 5.7% | 0% | **65%** | 5.9% |
| **Invalid** (0.5%) | 0% | 0.6% | 0% | 0% | 0.6% | 1.2% | **5%** | 1.2% |
| **No vote** (10.6%) | 7.8% | 8% | **2.2%** | **2.1%** | 8.2% | **18.1%** | **20%** | **21%** |
| **Total** | **100%** | **100%** | **100%** | **100%** | **100%** | **100%** | **100%** | **100%** |

*Table A8: Distribution of partisanship and reported vote choice according to GLES (N=1827).*

*Note:* Not including missing completions; Column percentages; overall totals of variables in parentheses.

## X. Robustness checks

### 1. Comparison of results using imputed and non-imputed personas

For our main analyses, we imputed missing values on any of the prompt variables for 377 respondents to obtain complete personas for prompting GPT. Most of these missing values concerned household income (248) and political ideology (114), with missing values for all other variables affecting less than 30 individuals, respectively, and no missing values for the key demographic indicators age and gender. As a robustness check, we re-prompt the LLM for these respondents using an adjusted prompt containing only the non-imputed, incomplete information (see Table A2b in Appendix II for an example prompt). We then merge these non-imputed observations and their GPT predictions with the a-priori complete cases to once again create a full sample and compare it to the main sample.

Relying on the non-imputed personas results in a higher overall share of missing predictions. While this share was 0.9% for the main sample after up to three trials, it is 2.3% for the total sample including the non-imputed personas. However, this difference has no impact on the substantive results. The aggregate vote shares estimated by GPT differ from the main results only by tenths of a percentage point, the only exception being the estimated share of non-voters now being two percentage points lower, 14.6 compared to 16.5%. The overall share and variance in matching vote choices between GPT predictions and GLES reports is the same. Using the non-imputed personas, both the matches by GLES party vote (recall) and the LLM's precision (matches by GPT prediction) per party are of similar magnitude, with minor changes in the rank order, as are the overall F1 scores per party.

As it lies in the nature of the non-imputed observations that they carry missing values on at least one predictor variable, we cannot compare the differential impact of the (missing) predictor variables on vote choice. However, we can employ a binary indicator for individuals with missing (sample with non-imputed personas) or imputed (main sample) values. Comparing multinomial regressions on GPT-predicted vote choice confirms the aggregate-level observation that for non-imputed personas, i.e. those with missing values, GPT is more likely to predict "Did not vote", an effect that is not present for the same people in the main sample including the imputed values. Considering that prompting GPT with the non-imputed personas led to a higher share of missing predictions to begin with, the effect of non-imputation on the completed predictions may even be underestimated. Distinguishing between the LLM's prediction of (individual) turnout and substantive vote choice could provide further insights into this matter. For an analysis of the effect of using imputed versus incomplete prompts on the successful completion of a prediction, see the next subchapter.

Taken together, these results suggest that GPT not only appears to be more likely to return a prediction, but also is more likely to predict a person voted when provided with more information about the person.

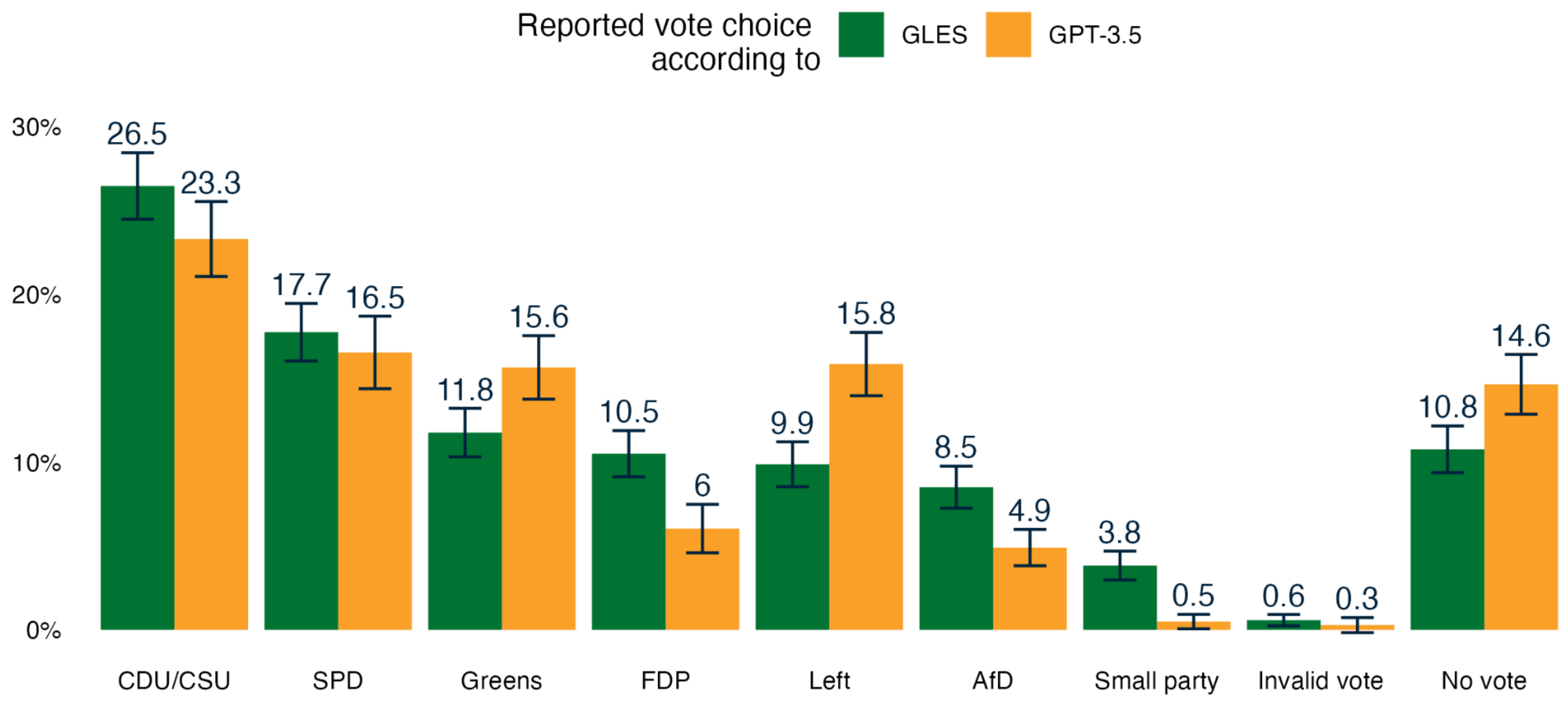

*Figure A2: Distribution of vote shares as estimated by GLES and GPT – robustness sample.*

| **Party** | **F1 Score** $\frac{2 \times precision \times recall}{precision + recall}$ |
|---|---|
| CDU/CSU | 0.63 |
| Greens | 0.52 |
| SPD | 0.52 |
| Left | 0.44 |
| FDP | 0.35 |
| AfD | 0.34 |
| No vote | 0.23 |
| Small party | 0.12 |
| Invalid | 0 |

*Table A9: Model accuracy (F1 scores) – robustness sample.*

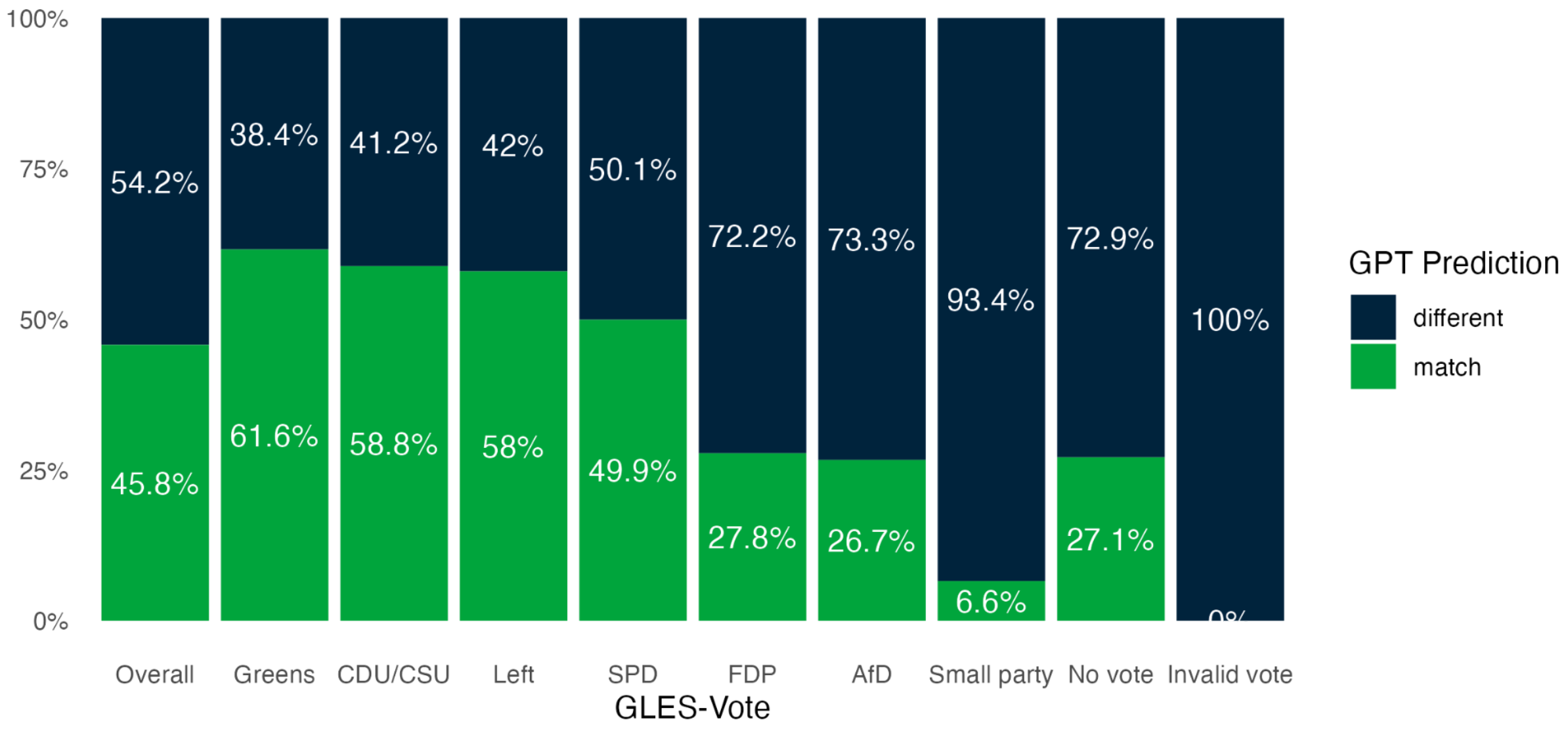

*Figure A3a: Share of party votes according to GLES that GPT correctly predicted (recall) – robustness sample.*

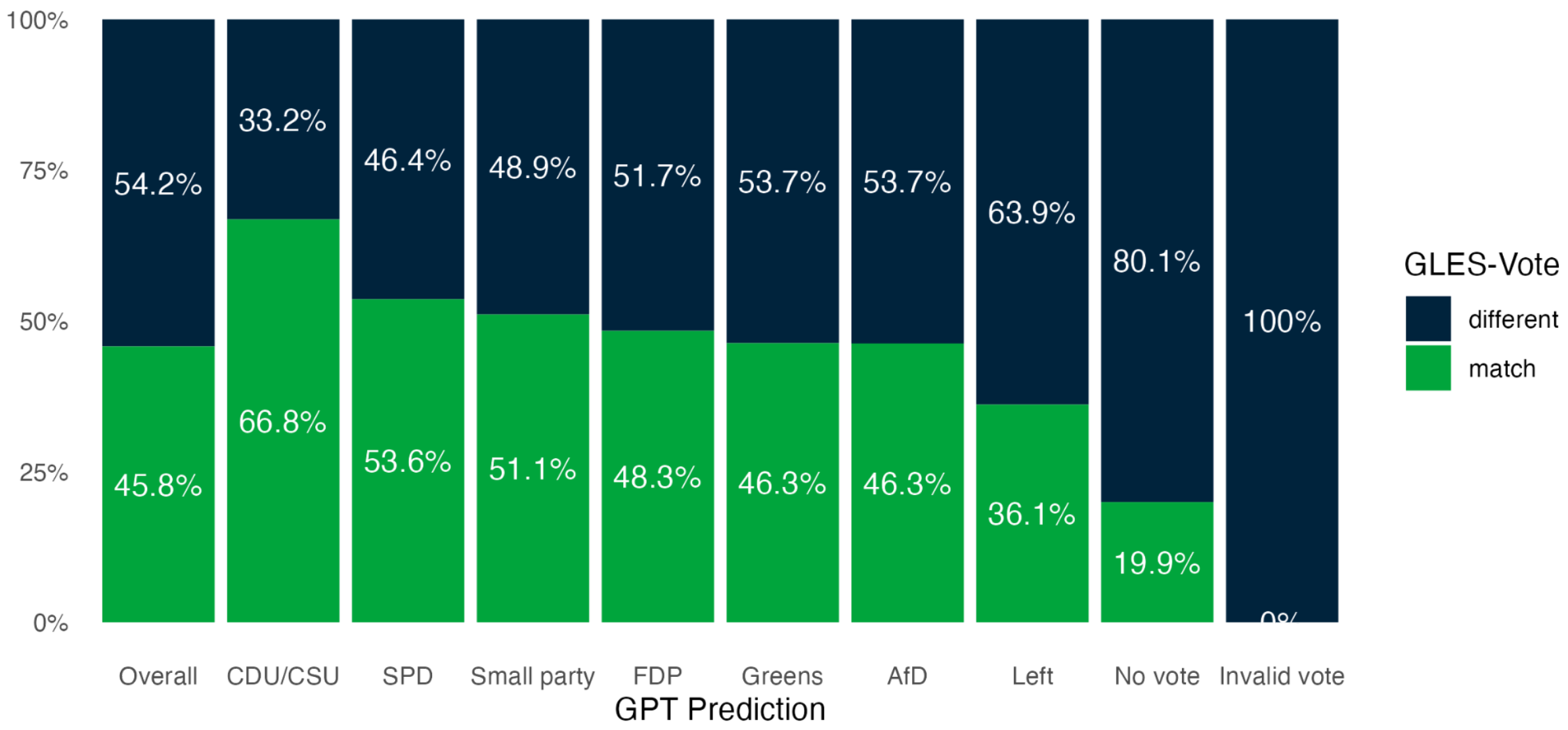

*Figure A3b: Share of GPT predictions that voted for the respective party (precision) – robustness sample.*

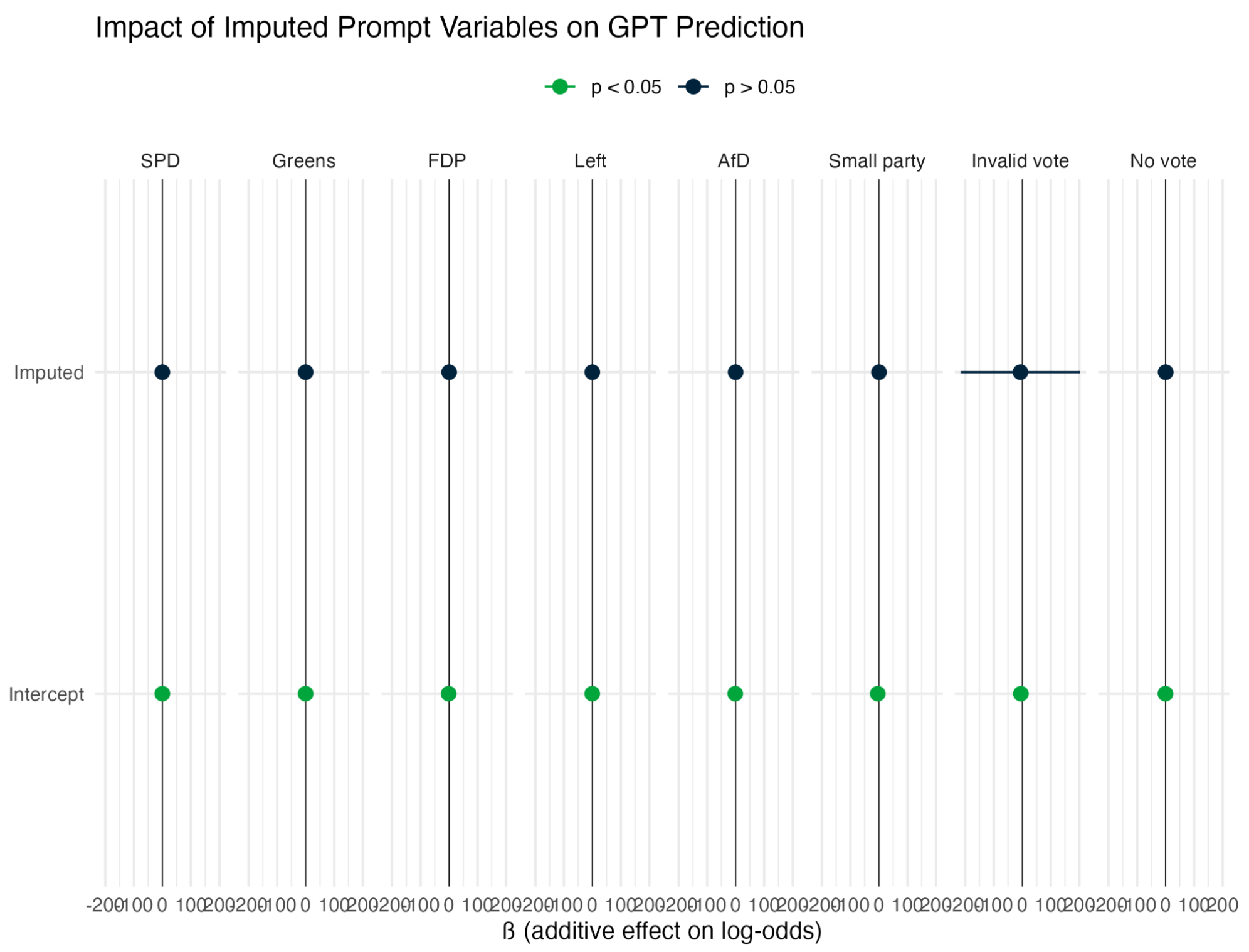

*Figure A4: Results of a logistic regression of indicator for imputed cases on GPT vote choice – main sample.*

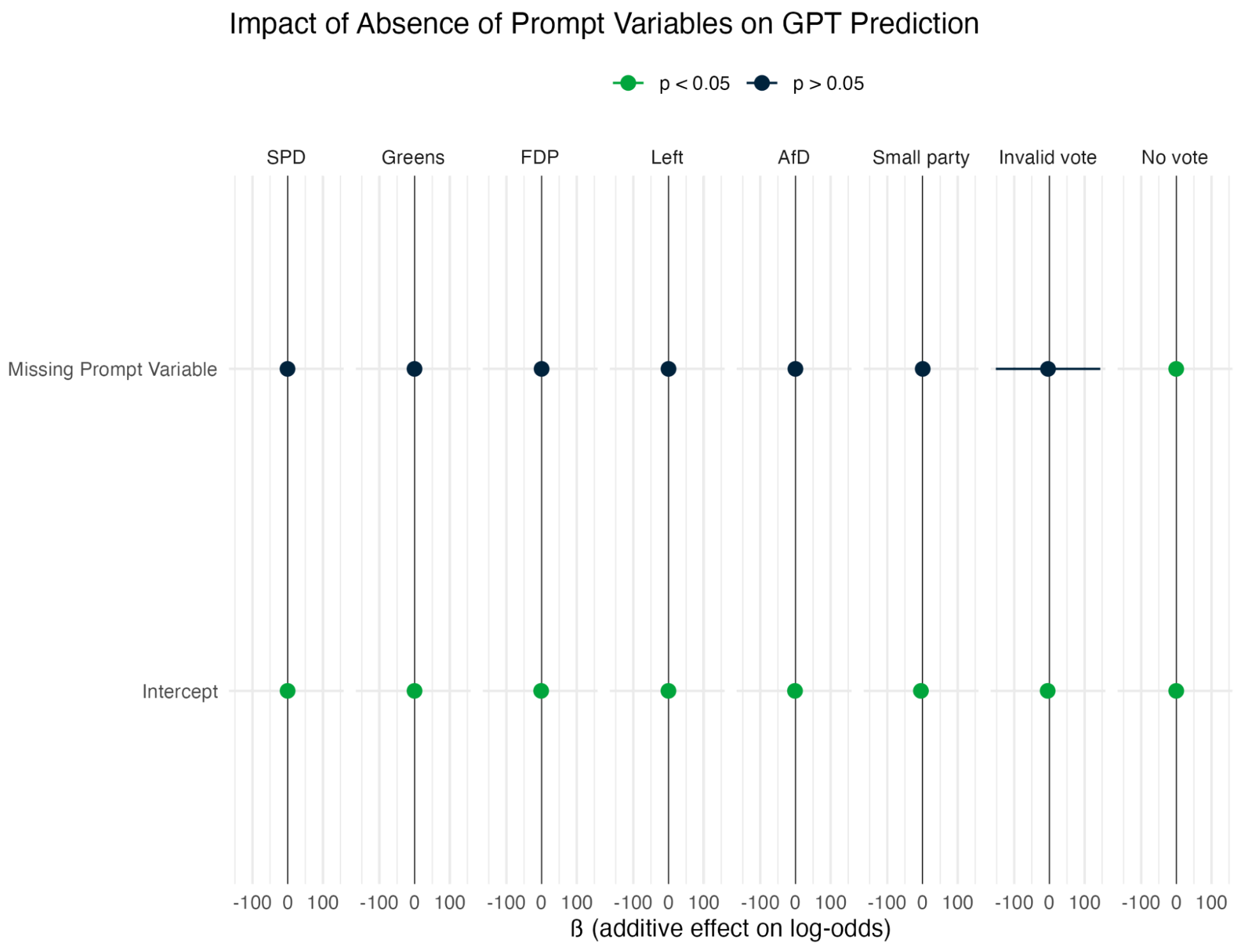

*Figure A5: Results of a logistic regression of indicator for cases with missing prompt variables on GPT vote choice – robustness sample.*

## 2. Analyses of missing completions

After re-sampling completions not containing an identifiable vote choice up to two times, 87 or 0.9% of completions remained without such information, corresponding to 78 individuals (4.3% of the sample), 9 of which have two missing predictions. On average, there are 15 missing predictions in each of the five iterations we ran for each persona, the fourth iteration being an outlier with 27 missing cases. For ensuring comparability across iterations, these individuals were excluded from the regression analyses. In the robustness sample, 222 completions (2.3%) do not contain a prediction, corresponding to 150 individuals (7.9% of the total sample), 114 of which have at least two and up to five missing predictions. Considering only the non-imputed personas in the robustness sample, 141 completions (7.5% of non-imputed personas) do not contain a prediction. At 43, the average number of missing predictions in each of the five iterations is much higher than in the main sample, and once again, the fourth iteration is an outlier with 50 missing cases. The differences in missing predictions between imputed and non-imputed personas support the notion that the full set of information is beneficial to a successful prediction – although the main results show that not all information is weighted equally in the outcome of the prediction.

In both the main and especially in the robustness sample, the shares of missing predictions increased over the course of the three rounds of sampling we performed for those completions that did not contain a vote choice in the previous round (see Table A4 in Appendix V). A descriptive analysis reveals that across samples, individuals for whom at least one GPT prediction was missing are on average younger than those with five predictions (47 vs. 52 years). The (binary) gender distribution among them is more balanced, while complete cases skew male. Those for whom a prediction was missing tend to be employed more often, resulting in fewer non-working individuals. The share of low-income individuals is lower among those with missing predictions, while that of medium-income individuals is higher. Ideologically, individuals with missing predictions tend to position themselves more in the middle, while there are more strongly-left and -right-leaning individuals among complete cases. Regarding partisanship, those with missing predictions largely do not identify with any party at all, much more so than complete cases. Notably, there are no Green or AfD partisans, but more small party partisans. Moreover, among those who do identify with a party, many more incomplete cases do so only weakly, compared to complete cases. The share of those supporting immigration is much lower among those with missing predictions. While these patterns are similar across the main and robustness samples, there are some differences regarding other prompting variables. For example, while the share of East German residents is higher among individuals with missing predictions in the main sample, the opposite is true in the sample containing the non-imputed cases. Although in both samples, incomplete cases tended to vote less for the CDU/CSU and SPD and more for the FDP, AfD (according to their survey response), only in the robustness sample was the share of those voting for the Greens lower and that of those voting for the AfD, small parties, or not voting at all, much higher among individuals with missing predictions. Interestingly, in the main sample (containing imputations), individuals with missing predictions are less likely to be imputed cases.

In sum, GPT appears to be more likely to make complete predictions for older, male, wealthier individuals who are ideologically unambiguous, strong (especially Green or AfD) partisans or voted for one of the bigger, centrist parties, and tend to support immigration. This echoes the bias observed in our main analyses, indicating that GPT tends to pick up on signals representing dominant or highly "visible" subgroups, while struggling with non-typical subgroups.

| | | **Main sample (including imputed values)** | | **Robustness sample (no imputed values)** | |
|---|---|---|---|---|---|
| Subgroup | | Individuals with 5 completions (N = 1827) | Individuals with at least one missing completion (N = 78) | Individuals with 5 completions (N = 1755) | Individuals with at least one missing completion (N = 150) |
| **Age** | average | 47 | 51 | 47 | 52 |
| **Gender** | male | 52.7% | 50% | 52.7% | 50.7% |
| | female | 47.3% | 50% | 47.3% | 49.3% |
| **Education** | no degree | 1.5% | 1.3% | 1.5% | 2% |
| | Hauptschule | 21.5% | 17.9% | 21.4% | 20.8% |
| | Realschule | 31.2% | 30.8% | 31.2% | 30.9% |
| | Abitur | 17.2% | 19.2% | 17.2% | 19.5% |
| | College | 28.5% | 30.8% | 28.8% | 26.8% |
| **Employment Status** | not working | 35.7% | 23.1% | 36.1% | 26% |
| | studying | 7.5% | 7.7% | 7.4% | 8.7% |
| | working | 56.8% | 69.2% | 56.5% | 65.3% |
| **Income** | low | 18.7% | 9% | 17.5% | 15.8% |
| | medium | 69.1% | 78.2% | 69.8% | 72.6% |
| | high | 12.2% | 12.8% | 12.7% | 11.6% |
| **Residence** | West | 67.9% | 62.8% | 67.4% | 70.7% |
| | East | 32.2% | 37.2% | 32.6% | 29.3% |
| **Religiosity** | not at all | 35.1% | 33.3% | 35.7% | 29.5% |
| | not very | 16.6% | 20.5% | 16.4% | 21.9% |
| | somewhat | 36% | 39.7% | 35.4% | 43.2% |
| | very | 12.2% | 6.4% | 12.4% | 5.5% |
| *cont.* | strongly left | 8.2% | 3.8% | 8.6% | 2.3% |
| | center-left | 26.1% | 24.4% | 26.7% | 20.8% |
| | in the middle | 55% | 66.7% | 53.5% | 70.8% |
| **Ideology** | center-right | 8.8% | 5.1% | 9.5% | 5.4% |
| | strongly right | 2% | 0% | 2% | 0.8% |
| **Party ID** | CDU/CSU | 28.9% | 15.4% | 30% | 9.5% |
| | SPD | 19% | 10.3% | 19.5% | 6.6% |
| | Greens | 10% | 0% | 10.6% | 0% |
| | FDP | 5.3% | 2.6% | 5.3% | 1.5% |
| | Left | 8.6% | 1.3% | 9% | 0.7% |
| | AfD | 4.5% | 0% | 4.7% | 0% |
| | Small party | 1.1% | 3.8% | 1.1% | 2.9% |
| | none | 22.4% | 66.7% | 19.8% | 78.8% |
| **Strength of Party ID** | none | 22.4% | 66.7% | 20.4% | 81.2% |
| | very weak | 0.6% | 66.7% | 0.6% | 0% |
| | rather weak | 2.5% | 0% | 2.5% | 2% |
| | moderate | 30.4% | 2.6% | 30.9% | 7.4% |
| | rather strong | 37.7% | 12.8% | 39% | 8.1% |
| | very strong | 6.4% | 16.7% | 6.6% | 1.3% |
| **Att. Inequality** | don’t act | 10.2% | 3.8% | 10.1% | 7.6% |
| | no opinion | 13% | 17.9% | 13.1% | 14.5% |
| | act | 76.8% | 78.2% | 76.7% | 77.9% |
| **Att. Immigration** | limit | 49.8% | 53.8% | 49.6% | 52.4% |
| | neither nor | 18.2% | 23.1% | 18.3% | 21% |
| | facilitate | 32% | 23.1% | 32.1% | 26.6% |
| *cont.* | CDU/CSU | 26.9% | 16.7% | 27.2% | 17.3% |
| | SPD | 18% | 11.5% | 18.1% | 13.3% |
| | Greens | 11.7% | 12.8% | 12% | 9.3% |
| | FDP | 10.3% | 15.4% | 10.4% | 11.3% |
| | Left | 9.9% | 9% | 9.7% | 11.3% |
| **Vote choice (GLES)** | AfD | 8.5% | 9% | 8.4% | 10% |
| | Small party | 3.6% | 9% | 3.6% | 6.7% |
| | Invalid | 0.5% | 1.3% | 0.6% | 6.7% |
| | No vote | 10.6% | 15.4% | 10% | 20% |
| **Imputed** | No | 80% | 92.3% | 100% | 100% |
| | Yes | 20% | 7.7% | 0% | 0% |

*Table A10: Distributions of subgroup characteristics by missing completion and imputation status (column percentages per variable; rounded values).*

## 3. Analyses without misclassified respondents

In some instances, GPT misclassified respondents as non-voters because it considered them ineligible to vote, mostly based on their age (all German citizens aged 18 and over are eligible to vote).

In 16 completions (14 unique respondents, 0.7% of the sample), the LLM wrongly stated that the respective respondent (between 58 and 94 years old) was too old to be eligible to vote and therefore did not vote.

Moreover, for 51 completions (44 unique respondents, 2.3% of the sample), GPT assumed respondents to be too young and therefore ineligible to vote. Only half of these cases (24, corresponding to 19 unique respondents) can be attributed to the fact that we did not specify that the prompt information referred to 2017, thereby possibly inducing GPT to adjust for the time difference of six years between the election and prompting. However, in the other half of cases (27 / 25 respondents), the respondent would have been over 18 in 2017 even if assuming the age information in the prompt referred to 2023, and in 16 of those cases, the respondent was between 30 and 94 years old.

Finally, GPT considered 17 cases (and respondents, 0.9% of the sample) ineligible to vote and therefore predicted "did not vote" without giving a specific reason for ineligibility.

However, all of these instances occurred at most twice per respondent, with the remaining three or four GPT completions per respondent containing a different vote choice. In total, 18 of the 73[8] respondents (25%) who GPT wrongly considered ineligible to vote and predicted to be non-voters because of this actually did not vote according to the GLES data.

Excluding these respondents from the analysis (along with those with less than five complete predictions) yields the following results: Even though the estimated vote shares change in absolute values for both the GLES and GPT data due to the omitted respondents, the relative differences remain the same. There continue to be no significant differences for the CDU/CSU and SPD. Even when excluding some of the individuals GPT had predicted to be non-voters, GPT significantly overestimates the share of Green, Left, and non-voters, and underestimates those of FDP, AfD, and small party voters. The F1 scores change slightly, but only on the second decimal, and do not substantially change the ranking.

[8] Two respondents were considered generally ineligible in one completion, too old in the other, hence being counted towards both groups but only once in the overall total.

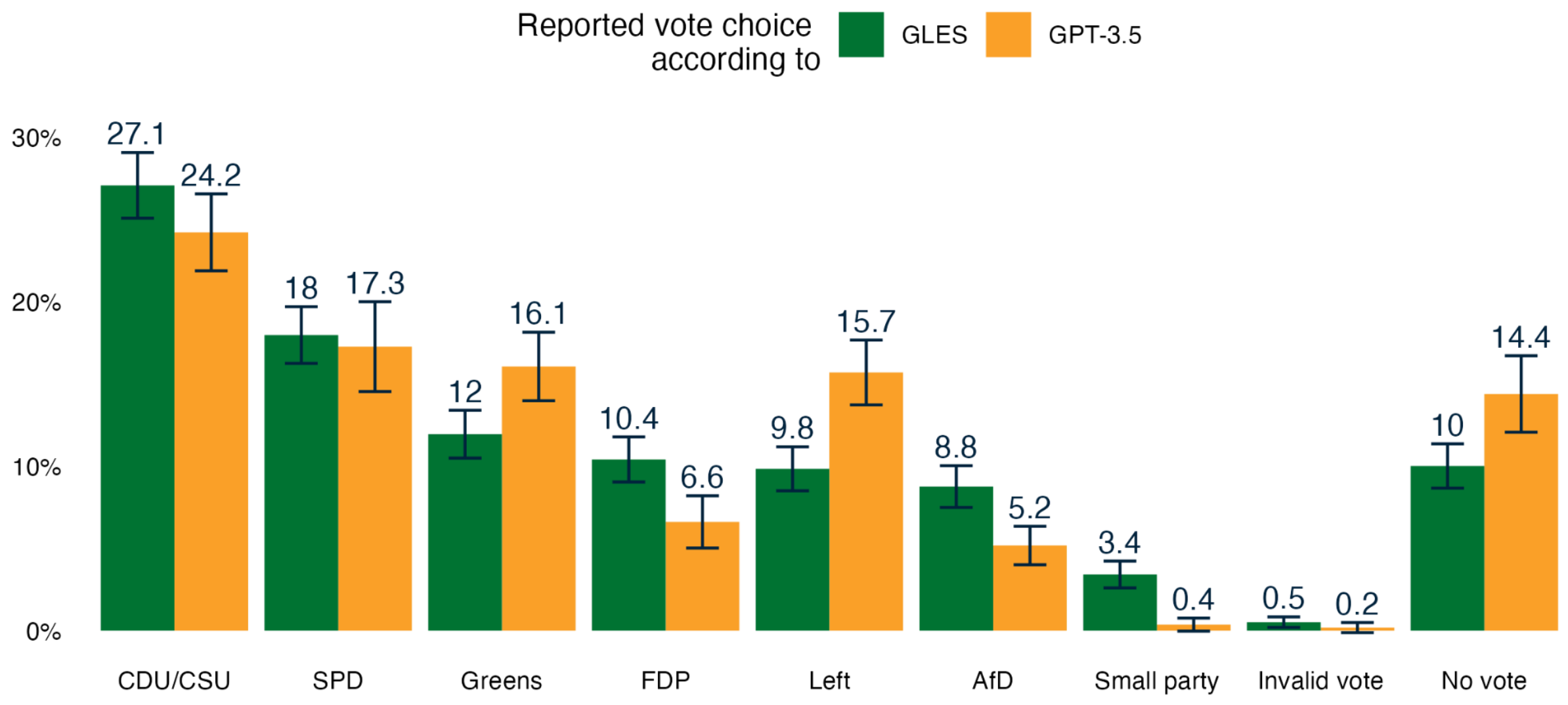

*Figure A6: Distribution of vote shares as estimated by GLES and GPT, excluding respondents considered ineligible.*

| **Party** | **F1 Score** $\frac{2 \times precision \times recall}{precision + recall}$ |
|---|---|
| CDU/CSU | 0.63 (+0.01) |
| Greens | 0.54 (+0.02) |
| SPD | 0.53 (+0.01) |
| Left | 0.46 (+0.01) |
| FDP | 0.36 (+ 0.02) |
| AfD | 0.35 (+ 0.02) |
| No vote | 0.22 (- 0.03) |
| Small party | 0.12 (+ 0.01) |
| Invalid | 0 |

*Table A11: Model accuracy (F1 scores), excluding respondents considered ineligible.*

## Data Availability

Data and code for the analyses conducted can be downloaded from https://github.com/leahvdh/Vox-Populi-Vox-AI